%% file: main.tex
\documentclass[twoside]{article}

\usepackage{caption}
\usepackage{amsmath}
\usepackage{multirow}
\usepackage{balance}
\usepackage{amssymb}
\usepackage{graphicx}  % for \resizebox
\usepackage{booktabs}  % for \toprule, \midrule, \bottomrule
\usepackage{siunitx}   % for \SI command
\usepackage{placeins}
\usepackage{pifont}% http://ctan.org/pkg/pifont
\usepackage{afterpage}
\usepackage[none]{hyphenat}
\usepackage[dvipsnames]{xcolor} % Load xcolor with extended named colors
\usepackage{hyperref}
\usepackage{pdflscape}
\usepackage{duckuments}
\hypersetup{
    colorlinks=false,       % Turns off colored text, uses colored boxes instead
    linkbordercolor=red,    % Color for normal internal links (red)
    citebordercolor=green,  % Color for citations (green)
    urlbordercolor=cyan     % Color for URL links (cyan as an example)
}

\usepackage[accepted]{claiunconf2023}
\usepackage[round]{natbib}

\newcommand{\cmark}{\ding{51}}%
\newcommand{\xmark}{\ding{55}}%

\begin{document}



% \textbf{Timeline}
% \begin{figure}[h]
% \includegraphics{unconf_deadline.png}
% \end{figure}

% \textbf{Discussion Question from Meeting}
% \begin{itemize}
%     \item What is the tradeoff in the selection of the frequency of snaphsots (how often you save the model)
%     \item Why is this analysis useful? What do we gain out of it?
%     \item How do we go about selecting the algorithms we plan to analyze?
% \end{itemize}
% Redefine the \and macro

\newpage
\twocolumn[

\CLAIUnconftitle{Examining Changes in Internal Representations of Continual Learning Models Through Tensor Decomposition}

\CLAIUnconfauthor{
  Nishant Suresh Aswani\textsuperscript{*, \textdagger} \unskip\enspace\enspace
  Amira Guesmi\textsuperscript{\textdagger} \unskip\enspace\enspace
  Muhammad Abdullah Hanif\textsuperscript{\textdagger} \unskip\enspace\enspace
  Muhammad Shafique\textsuperscript{*, \textdagger} 
}

\CLAIUnconfaddress{
    \textsuperscript{*}Department of Computer Science and Engineering, New York University Tandon, Brooklyn, USA 
    \\ 
    \textsuperscript{\textdagger}eBrain Lab, Division of Engineering, New York University Abu Dhabi, Abu Dhabi, UAE
}

]
\begin{abstract}
%\textcolor{red}{To Do}

Continual learning (CL) has spurred the development of several methods aimed at consolidating previous knowledge across sequential learning. Yet, the evaluations of these methods have primarily focused on the final output, such as changes in the accuracy of predicted classes, overlooking the issue of representational forgetting within the model. 
%In this paper, we propose a novel representation-based evaluation framework to gain deeper insights into the phenomenon of representational forgetting in CL. Our framework centers on systematically formulating various three-dimensional tensors and employing Tensor Component Analysis (TCA) to model learning dynamics over time. This approach is expected to allow explanations of how the model's representations evolve, as the model encounters new tasks, shedding light on the forgetting or preservation of previously learned information.  
%In this paper, we present a comprehensive study that delves into the dynamics of continual learning, shedding light on key insights regarding neural adaptation, filter evolution, and model representations. %Our framework encompasses various components, including neural activations collection by performing model snapshots at defined intervals during the continual learning process, tensor construction to capture relevant information, and 
%Our framework centers on systematically formulating various three-dimensional tensors and employing Tensor Component Analysis (TCA) to reveal patterns within hidden representation internal state changes. TCA helps identify significant components and similarities among different tasks, inputs, and model snapshots, leading to valuable insights into the continual learning process.
In this paper, we propose a novel representation-based evaluation framework for CL models.
This approach involves gathering internal representations from throughout the continual learning process and formulating three-dimensional tensors. The tensors are formed by stacking representations, such as layer activations, generated from several inputs and model `snapshots', throughout the learning process.
By conducting tensor component analysis (TCA), we aim to uncover meaningful patterns about how the internal representations evolve, expecting to highlight the merits or shortcomings of examined CL strategies. We conduct our analyses across different model architectures and importance-based continual learning strategies, with a curated task selection. While the results of our approach mirror the difference in performance of various CL strategies, we found that our methodology did not directly highlight specialized clusters of neurons, nor provide an immediate understanding the evolution of filters. We believe a scaled down version of our approach will provide insight into the benefits and pitfalls of using TCA to study continual learning dynamics.
% TCA is a powerful technique that helps identify significant components and commonalities across time, tasks, and inputs. It allows us to gain valuable insights into how the model's internal representations evolve, adapt, and specialize over time, providing a deeper understanding of the continual learning dynamics.
\end{abstract}

\input{v4/1_introduction}
\input{v4/2_related}
\input{v4/3_proposed}

\input{v4/4_experiment}
\input{v4/5_changes}
\input{v4/6_results}
\input{v4/9_future_work}
\input{v4/7_conclusion}

%% FOR CAMERA READY VERSION
% \input{8_acknowledgements}
%% FOR CAMERA READY VERSION

% \newpage 

%===============================================================================

% \newpage 

% \newpage 

%\balance
\clearpage
% \subsubsection*{References}
% \newpage
% \balance
\bibliography{refs}{}

% If you have textual supplementary material
\newpage 
\appendix
\renewcommand{\thefigure}{\thesection.\arabic{figure}}
\renewcommand{\thetable}{\thesection.\arabic{table}}
\onecolumn
\input{v4/supplementary_a}
\input{v4/supplementary_b}

\newpage
\input{v4/supplementary_c}

\input{v4/supplementary_d}

\vfill

\end{document}

%% file: v4/1_introduction.tex
\section{INTRODUCTION}
% \fixme{All references to the CIFAR datasets must be in the format "Split-CIFAR-X-(Y)"}

Learning is a core capability for all intelligent agents. While biological agents acquire new knowledge and adapt by building upon their prior learning experiences, artificial agents follow a more static and meticulously curated learning process. 
%Biological agents are capable of more effectively learning new patterns throughout their lifetime, building atop previous learning experiences. Artificial agents, on the other hand, experience a static, meticulously curated learning procedure. 
When challenged to encounter new concepts, a setting familiar to biological agents, machine learning models suffer from \textit{catastrophic forgetting}, demonstrating poor performance when attempting to recall old patterns \citep{schlimmer1986case, ring1997child}. As continual learning (CL) research matures in tackling this challenge, solutions will look increasingly composite, likely layering multiple mechanisms to mitigate forgetting and achieve additional goals \citep{kudithipudi2022biological}. %This complexity necessitates an architecture and strategy agnostic explainability tool aimed at understanding how CL methods help models learn new tasks and prevent forgetting on previous ones. 
The growing complexity underscores the need for an architecture and strategy agnostic explainability tool designed to shed light on how CL methods enable models to acquire new tasks while preventing the forgetting of previously learned ones.
The insights gained from studying how representations change when experiencing catastrophic forgetting has previously facilitated the advancement of new methods. Thus, better understanding how existing methods update model parameters over time %may
holds the potential to inspire the development of even more effective learning strategies. 

%===================================================
\begin{figure*}[!t]
%\vspace{.3in}
%\centerline{\fbox{This figure intentionally left non-blank}} 
\centering
\includegraphics[width=\textwidth]{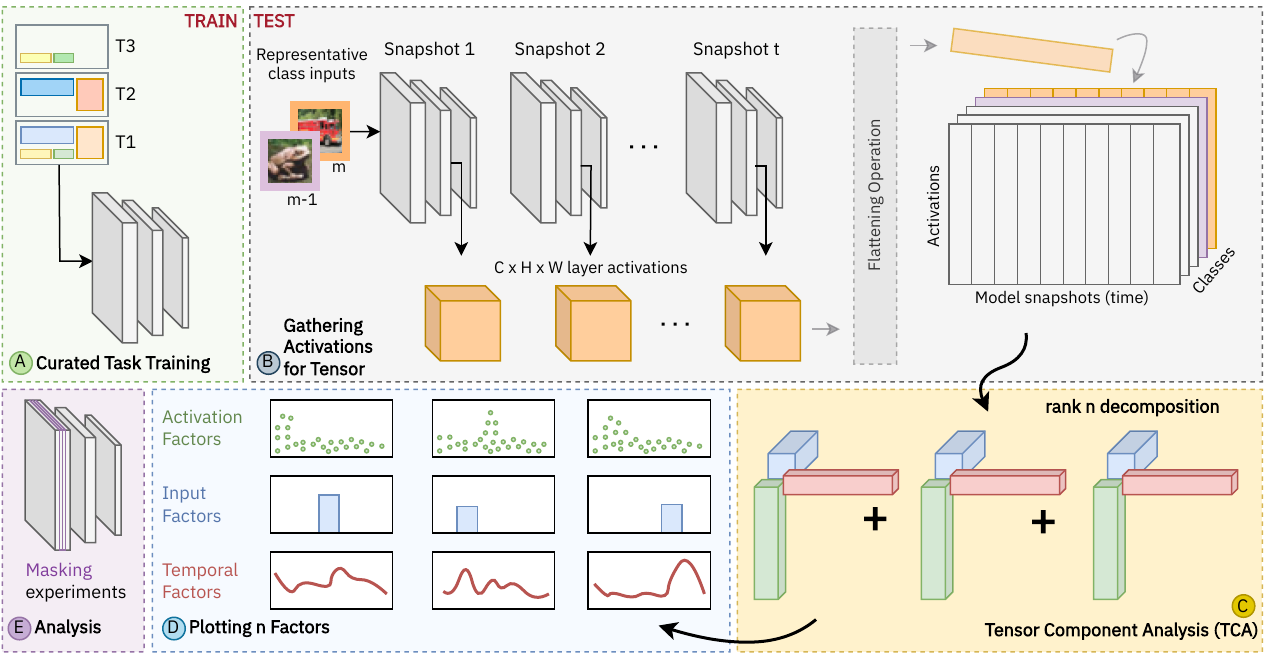}
%\vspace{.3in}
\caption{Overview of the proposed methodology. (A) We train a model with a CL method with curated tasks. (B) We assume access to `snapshots' of the model taken throughout training on all tasks. At inference time, for a given dataset class, we feed its corresponding representative inputs into each of the snapshots and, for a layer of choice, gather the resulting activation tensors. We flatten the activations into a vector and stack them for all snapshots to obtain a matrix. We repeat this process for all inputs and stack the matrices to obtain a tensor. (C) We conduct tensor component analysis (TCA) with rank $n$ to obtain $n$ components. (D) We plot $n$ components, each of which consists of three factors: factors selecting for certain activations (green), factors selecting for certain inputs (blue), and factors describing a temporal activity (red). (E) We conduct model masking experiments to verify our observations.}
\label{methodology}
\end{figure*}

To execute our study, we propose a novel framework that allows us to explore internal state changes during the continual learning process. Our framework centers on creating a data representation that captures the model's internal representations across different tasks over time. By leveraging Tensor Component Analysis (TCA) \citep{williams2018unsupervised}, a technique for three-dimensional tensor decomposition, we expect to uncover patterns concerning internal representations across time. The proposed methodology, as depicted in Figure \ref{methodology}, provides an overview of our study in understanding how representations evolve during CL. %Our framework enables the investigation of representational changes as new tasks are encountered, shedding light on the adaptability and potential forgetting mechanisms within the model.

% In summary, the contributions and the key insights that can be derived from this analysis are:
\textbf{In summary, the contributions and the key insights that can be derived from this analysis are:}
\begin{itemize}
    % \item We propose a novel framework for analyzing CL algorithms. The framework provides a systematic approach to analyze models over time, yielding valuable insights into how different continual learning strategies impact a model's learned representations across tasks and throughout task training. 
    \item We leverage TCA, an unsupervised method for analyzing three-dimensional tensors, to extract patterns about how model representations evolve in a continual learning setting. TCA results in a data representation that potentially captures the model's dynamics for multiple inputs over time. To the best of our knowledge, we are the first to leverage TCA in continual learning interpretability.
    \item We systematically compare the performance, and analyze the internal representations of several parameter-based methods (and their combinations with replay) across convolutional neural networks (CNNs), vision transformers, and a CNN-transformer hybrid architecture. In addition, we conduct these experiments with a curated set of tasks to study how internal representations change based on the feature similarities of tasks.
    % \item We explore the evolution of hidden representations in the model as it learns new tasks, the study can reveal how the model's internal representations adapt and change over time which can provide crucial information on its adaptability and ability to generalize to novel data distributions. 
    %Understanding how the model accommodates new tasks can provide crucial information on its adaptability and ability to generalize to novel data distributions.
    \item We investigate neuron tracking and filter evolution which can shed light on the neural plasticity that occurs during continual learning. We plan to gain insights on which neurons or filters are more flexible and likely to change their responses over time, and which ones remain stable. 
    %This knowledge is essential for understanding the underlying mechanisms of continual learning and how the model accommodates new information without catastrophic forgetting.
    %\item We analyze how changes in neural activations impact the model's performance, the study can establish correlations between representation dynamics and performance fluctuations. It can provide insights into how changes in hidden representations influence the model's ability to learn and retain knowledge across multiple tasks.
    %\item The comparison of importance-based regularization methods, replay-based approaches, and cumulative baseline methods can elucidate how different strategies influence the model's internal representations. This knowledge can inform the development of more effective regularization techniques to improve continual learning performance.
\end{itemize}

\textbf{Summary of Hypotheses:}

\noindent \textbf{Hypothesis Set 1.} Do importance-based regularization methods lead to the emergence of `specialized' neurons for specific tasks? Does adding replay reinforce this specialization? Is there a consistent behavior across several model architectures? By tracking activations throughout training with various CL strategies, we will study the temporal patterns of these activations.
% By tracking neurons throughout training with parameter regularization, we aim to uncover patterns in neuron specialization and understand how these specialized neurons evolve over time.

\noindent \textbf{Hypothesis Set 2.} Do CNN filters or transformer features within the same layer exhibit different update patterns throughout training? How do these update patterns compare by CL strategy and model architecture? By tracking what CNN filters and transformer features select, we will examine how filters/features %change over task-incremental learning.
evolve throughout the task-incremental learning.

% We will investigate whether certain filters/features update earlier or later in training and how these updating patterns vary across different regularization methods. By optimizing images to activate specific filters/features, we aim to study filter evolution and its implications for model learning.

% \noindent\textbf{Hypothesis 3:}  {\color{red} TODO}

% \noindent\textbf{Hypothesis 3:} Multi-head and single-head models learn different internal representations when trained on the same tasks. We explore how the hidden representations of these models differ and whether there are significant variations in the learned features across different depths of the model. By comparing multi-head and single-head model representations, we aim to understand the impact of model architecture on continual learning performance.

%These hypotheses collectively form a comprehensive study aimed at understanding the dynamics of continual learning, including neuron specialization, filter evolution, and the role of model architecture. The insights gained from these investigations can contribute to the development of more efficient and adaptive continual learning algorithms, leading to improved model performance and adaptability in dynamic environments.

%% file: v4/2_related.tex
\section{BACKGROUND \& RELATED WORK}

\subsection{Techniques for Examining Representation Quality}

By examining representations, one may assess how well a model's learned features generalize across tasks and the degree to which they preserve task-specific information.
% By scrutinizing representations, it becomes possible to gauge the extent to which a model's learned features generalize across tasks and the degree to which they preserve task-specific information.
Strong representations may enable better knowledge transfer and improved adaptation to new tasks. In this section, we explore several works studying representations in the context of CL. These studies employ various metrics and experimental scenarios to evaluate the quality of learned representations during training.

% to gain an understanding of how representations evolve and adapt during continual learning.

\cite{ramasesh2020anatomy} studied the impact of catastrophic forgetting on hidden representations by quantitatively comparing the layer representations before and after sequentially training on a second task. To measure the similarity between representations across multiple layers of the model, they utilized the centered kernel alignment (CKA) technique. They also investigated how the semantic similarity between tasks affected forgetting, obtaining nuanced results suggesting that forgetting is maximized when sequential tasks have intermediate rather than low or high similarity.

% CKA is employed to assess how similar the representations are across different layers.
%They employed representational similarity measures \cite{kornblith2019similarity, raghu2017svcca} and conducted layer freezing and layer reset experiments. 
%The results of their investigation indicated that deeper layers within the neural network were the main contributors to the phenomenon of forgetting. These deeper layers underwent the most significant changes during sequential training, leading to the observed effects of catastrophic forgetting. 

% \cite{Murata2020} developed a re-training-based strategy to evaluate the representation in CL on past task and examine the performance of lower layer representations. Findings from the study indicate that there is a non-negligible amount of representational forgetting that is observed at the shallow layers of a deep neural network model and the accuracy of task learning in the presence of representational forgetting depends on the layer depth where the forgetting occurs. 
 
\cite{davari2022probing} evaluated the quality of representations using Linear Probes (LP), which involves training a linear classifier on the fixed representations for each task. The accuracy of this linear classifier on the task's test set served as a metric to assess the representation quality for that particular task.
To measure forgetting, the authors measured the difference in linear probe performance before and after introducing a new task. 
% This performance change acted as a surrogate measure for the amount of forgetting observed in the representations. 
% Comparing linear probe accuracies allowed the authors to assess the similarity of representations across different depths and models from various tasks, providing insights into the representation learning process.

\cite{zhang2022feature} created a synthetic dataset in which alternating (odd and even) tasks shared the same low-level features while each task simultaneously contained unique, high-level features. By having access to the `ground-truth' features, the authors examined whether the model was making progress towards learning the features encoded in the inputs.
% made it possible to follow the evolution of the features and to understand how they are learned by the model during the CL. 
% To simulate a real-world scenario, where CL tasks are often related and share common features, they generate a distribution of regression tasks. These tasks share a common set of underlying features, reflecting the interrelated nature of tasks in the real world. Additionally, each task within this distribution also possesses its unique high-level features and outputs, representing the distinct characteristics and objectives of individual tasks. To study the learning of features, they calculated the correlation between the network activations and the ground-truth features for each network layer. 
%Their results indicated that consistently encountering the shared feature in all even-numbered tasks prevented the model from completely learning the shared feature present in the odd-numbered tasks. The same was true for learning the shared feature in even-numbered tasks.
Their findings revealed that when the shared feature was consistently encountered in all even-numbered tasks, it prevented the model from fully learning the shared feature present in the odd-numbered tasks. The same held true for learning the shared feature in even-numbered tasks.
% Experiments showed that although the accuracy using the continually trained classifier degrades heavily due to catastrophic forgetting, the learned knowledge in the final feature space is well maintained.
%To study the learning of shared features assumed linear, they used correlation between the network activations and the ground-truth features for each network layer. For non-linear features representing the distinctive high-level features, they used a fixed two-layer rectifier network to generate the shared features making them piece-wise linear

\cite{hess2023knowledge} introduced a task exclusion comparison, hypothesizing that if a model has trained on a task and retained knowledge related to that task, then it should exhibit a more robust representation of that task compared to a model that has never encountered the task. To test this, they compared the linear probe accuracy of models that have encountered a task with those that have had the same task deliberately excluded. Their results indicate that continually trained models typically forget task specific knowledge quickly, contrary to what was presented by \cite{davari2022probing}.
%Based on this, they compared the linear probe accuracy of models that have encountered a task with a model that has had the task excluded. Their results indicate that continually trained models typically forget task specific knowledge quickly, contrary to what was presented by \cite{davari2022probing}.
%findings: A model that was not trained on a specific task can achieve nearly equivalent linear probe accuracy, suggesting that a significant portion of the task-specific information in the representation has been forgotten.
%%%

Across most works studying layer representations in the context of continual learning, CKA and linear probes make a recurring appearance as tools to measure the similarity of hidden layer activation patterns in neural networks. %While these metrics may provide a general insight into how representations compare, the sometimes conflicting results obtained by existing literature using similar experimental settings highlight the need for a different approach to studying CL dynamics.
Although these metrics offer a broad understanding of representation comparisons, the occasional conflicting results among existing literature using similar experimental setups highlights the need for a different approach to studying CL dynamics.
Recent representation similarity work has even highlighted the sensitivity of CKA to outliers \citep{davari2022reliability}, urging researchers to ensemble similarity measurement using several techniques. 
% One commonly suggested technique for comparing network representations is the Orthogonal Procrustes distance, as it performs more consistently across a variety of tasks, as demonstrated by \cite{ding2021grounding}.

Nevertheless, most analyses of network representations in the context of CL limit themselves to similarity measurements of network representations pre and post-task training. We believe there is an opportunity to move away from drawing conclusions about network representations from ambiguous similarity measures when attempting to understand catastrophic forgetting. Instead, we suggest that it is possible to take advantage of unsupervised tensor decomposition to study how internal representations evolve across time for several inputs.

\subsection{Tensor Component Anlysis (TCA) for Exploring Learning Dynamics}
% TCA is fit only using iterative optimization algorithms. 
%tensor component analysis original paper \\
% https://www.cell.com/neuron/pdfExtended/S0896-6273(18)30387-8

Tensor component analysis (TCA), also known as canonical polyadic (CP) decomposition, is a tensor decomposition technique to identify variability across the three axes of a tensor \citep{carroll1970analysis, harshman1970foundations}. While it is a  dimensionality reduction technique similar to Principal Component Analysis (PCA), TCA differs by extending the decomposition to an additional axis. Further, unlike PCA, the factors obtained by TCA are not necessarily orthogonal, allowing for an expression of more natural patterns. For PCA, unless the features present in the data are naturally orthogonal to each other, the components recovered cannot be interpreted as those underlying features. In addition, PCA yields multiple solutions that can be employed to reconstruct the original data, known as the rotation problem \citep{williams2018unsupervised}. In practice, TCA does not face this limitation. %While TCA is not completely free of alternative solutions, the number of solutions is more limited and the solution to non-negative TCA is typically unique \citep{kruskal1977three, adali2022reproducibility}. Through various contexts in neural statistics, \cite{williams2018unsupervised} demonstrate that TCA is a unsupervised technique capable of demixing neural data and providing interpretable results corresponding to agent behavior and learning.
 While TCA is not entirely immune to alternative solutions, the number of potential solutions is more constrained, and in the case of non-negative TCA, the solution is typically unique \citep{kruskal1977three, adali2022reproducibility}. In various contexts within neural statistics, \cite{williams2018unsupervised} demonstrated that TCA serves as an unsupervised technique capable of demixing neural data and providing interpretable results corresponding to agent behavior and learning patterns.

% The main goal of TCA is to find a low-dimensional representation of tensor data while preserving its essential information. By reducing the dimensionality, TCA aims to extract meaningful patterns or features from complex data structures, making it easier to interpret and analyze.
%tca used for study in visual cortex \\
% https://www.sciencedirect.com/science/article/pii/S0960982222002500?via%3Dihub

While \cite{williams2018unsupervised} discussed TCA as a method to enhance neural data analyses, such that it describes trial-to-trial variability and avoids a trial-averaging step, \cite{mcguire2022visual} took advantage of TCA in a slightly different fashion. Studying the behavior of neurons in the postrhinal cortex (POR) of mice, the authors attempted to identify neuron population clusters, as well as the clusters' responses over time to various cues displayed to the mice as they sequentially learned two tasks. % In their research, the authors investigate the activity patterns of neurons within the postrhinal cortex (POR) of mice. Their objective is to discern distinct clusters within the neuron population and track how these clusters' responses evolve over time in response to different cues presented to the mice. Notably, the mice are sequentially trained on two tasks, and the study aims to uncover how the neuron population adapts its behavior during this learning process.
Rather than focusing on trial-to-trial variability, the study focused on identifying within-stage and across-stage dynamics for multiple cues. In this context, `stage' refers to the learning stages across two tasks. Through TCA, the researchers observed a `division of labor' in the neural populations, showing that certain clusters of neurons are specialized to activate in response to particular cues. Moreover, they were able to capture how these clusters vary their response across time.

% TCA has been used in the context of the study of the visual cortex in neuroscience. It can be applied to analyze multi-dimensional neural data \citep{mcguire2022visual}. This data may consist of responses from multiple neurons or brain regions to different visual stimuli, which naturally form higher-order tensor structures. TCA is particularly suitable for this type of data as it can handle and extract patterns from high-dimensional and multi-modal datasets.

% TCA was utilized as a data analysis technique to account for changes in both magnitude and shape of the cue response across different stages of learning and for differences in cue preference across neurons.

\cite{dyballa2023population} used TCA as part of their analyses to organize neurons into a `manifold' to study how the neurons they measured lie with respect to each other in a lower dimensional space. Generating these manifolds for biological and artificial neurons allowed the authors to compare the primary visual cortex (V1) to a CNN by studying the differences in how neurons are encoded within their manifolds. To accomplish this, the authors first employed TCA to obtain components which formed a lower dimensional space. Then, this component space was transformed into a manifold using the IAN similarity kernel and diffusion maps. The authors claimed that their analyses allowed them to organize neurons accounting for both stimulus features and temporal response, rather than the conventional approach of solely focusing on stimulus selectivity. In essence, the authors highlighted the ability of TCA to unlock an additional dimension, as well as provide outputs that can be fed into further analyses for network comparisons.

% The authors claim that the process of building these manifolds reveals that the manifold topology of CNN neurons is distinct from the manifold topology of neurons they measured in the primary visual cortex (V1).

%% file: v4/3_proposed.tex
\section{PROPOSED METHODOLOGY} %DYNAMICS OF CONTINUAL LEARNING
Through tensor component analysis (TCA), we plan to investigate how the internal representations of CL models evolve as the models train on incoming tasks. As suggested earlier, TCA unlocks an opportunity to ask questions about learning patterns that might emerge in the temporal dimension while looking at several inputs. 
The main questions that we will explore in this investigation are:
\begin{itemize}
    \itemsep0em 
    \item Are continual learning strategies able to elicit neuron specialization as they encounter new tasks? 
    \item How do convolutional filters and transformer features shift over time when encountering new data in the continual learning setting? 
\end{itemize}

For this exploration, we plan to use combinations of several CL strategies and model architectures. We look towards insights provided by simple explainability techniques to construct an understanding of continual learning dynamics.
% We seek to harness the insights offered by straightforward explainability techniques, using them as building blocks to construct a comprehensive understanding of the dynamics at play in continual learning.

\subsection{Overview}
Our proposed framework (see Figure \ref{methodology}) involves sequentially training a model on a stream of curated tasks (see Section \ref{sec:protocol}) and analyzing its representations. To achieve this, we adopt the following steps: 

\noindent \textbf{Saving Model Snapshots} During the continual training process, we periodically save snapshots of the model at defined intervals.
% These snapshots capture the state of the model at different stages of learning on the stream of tasks. 
For all tasks, we save multiple snapshots of the model while it is still learning the task, in order to conduct within-task and across-task analyses.

\noindent \textbf{Probing Model Snapshots} 
% After training the model and obtaining snapshots at specific intervals, 
Then, we probe each model snapshot. % For example, feeding inputs to the model and recording the layer from a selected layer within the model. 
This process involves feeding data to the model and recording, for example, the activations from a designated layer within the model.
Activations represent the outputs of a layer in response to the given inputs. %The collected neural activations are organized into a three-dimensional tensor. However, as we describe in later sections, we do not limit ourselves to activations.
The neural activations collected during this process are structured into a three-dimensional tensor. As we will later elaborate, our analysis extends beyond neural activations.

  \begin{table*}[!t]
    \renewcommand{\arraystretch}{1.05}
    \caption{\textbf{A Selection of Strategies Based on Parameter Importance}} 
    \label{table:strategies}
    \begin{center}
      \begin{tabular}{p{3.0cm} p{6.0cm} p{1.0cm} p{4.0cm}}
        
        \textbf{Approach} & \textbf{Strategy} & \textbf{+ ER?} & \textbf{Ref.} \\
        \hline \\[-1.5ex]
        
        \multirow{2}{*}{\parbox[t]{2.8cm}{Baselines}} & Naive             & \xmark  & \\
                                                     & Cumulative        & \xmark  & \\
        \\[-1ex] \hline \\[-1ex]
        
        \multirow{1}{*}{\parbox[t]{2.8cm}{Replay}}    & Experience Replay (ER) & \xmark & \\
        \\[-1ex] \hline \\[-1ex]
        
        \multirow{2}{*}{\parbox[t]{2.8cm}{Importance-Based \\ Regularization}}  
                                                                              & Elastic Weight Consolidation (EWC) & \cmark & \cite{kirkpatrick2017overcoming} \\
                                                                              & Memory Aware Synapses (MAS)  & \cmark & \cite{aljundi2018memory} \\
                                                                              % & Synaptic Intelligence (SI) & \cmark & \cite{zenke2017synaptic} \\
                                                                              % & Adaptive Group Sparse Regularization (AGS-CL) & \cmark & \cite{jung2020ags} \\
        \\[-1ex] \hline \\[-1ex]
        
        \multirow{2}{*}{\parbox[t]{2.8cm}{Importance-Based \\ Subnetworking}} & Relevance Mapping Networks (RMN) & \cmark & \cite{kaushik2021rmn} \\
                                                                             & Winning Subnetworks (WS) & \cmark & \cite{kang2022wsn} \\
                                    
      \end{tabular}
    \end{center}
    \renewcommand{\arraystretch}{1}
  \end{table*}

\noindent \textbf{Tensor Component Analysis} 
% This tensor contains dimensions corresponding to different aspects of the data. 
% One dimension might represent time (capturing the different stages of learning through the model snapshots), another dimension represents the inputs used to generate the activations, and the third dimension represents neural activations. 
%By decomposing this tensor with TCA, we expect to obtain interpretable components that, for example, describe how certain neural activations evolve during the learning process. This kind of analysis can offer insights into how the model's performance may be impacted by the changes in activations over time.
Through tensor decomposition, we hypothesize deriving interpretable components. These components can, for instance, provide descriptions of how specific neural activations evolve throughout the learning process. Such analysis holds the potential to offer valuable insights into how changes in activations over time might correspond to the model's performance.

\subsection{Problem Setup}
We begin with the standard setup of a CL problem, as described by \citet{hess2023knowledge}, where we assume an incoming stream of classification tasks $\mathcal{T} = \{\mathcal{T}_1, \mathcal{T}_2, ..., \mathcal{T}_i, ..., \mathcal{T}_t\}$. Each task consists of a set of images $X_i$ with their corresponding class labels $Y_i$. In the case of a multi-head classifier, the model also has access to the unique task identifier $i$. Section \ref{sec:protocol} discusses how we curate our tasks to aid the analysis. For a chosen CL strategy, we train a model $f_\theta$ sequentially on these tasks. For the purposes of a temporal analysis, we save a snapshot of the model parameters $\theta_{i,e}$ at defined intervals, where $i,e$ refers to the parameters of a model learning a task $\mathcal{T}_i$ after having completed training epoch $e$. For instance, if we train a model for 50 epochs on each of 3 tasks and take snapshots at a frequency of 10 epochs, we would obtain a set of snapshots $\theta = \{\theta_{1,10}, \theta_{1,20}, ..., \theta_{3,40}, \theta_{3,50}\}$.

\subsection{Tensor Formulation Details}

One way to probe a model snapshot $\theta_{i,e}$ is with an input $m$ to obtain network activations $A^m_{i,e}$ from a desired model layer. We can repeatedly probe that layer with $n$ different inputs, to obtain $n$ different activations $A_{i,e} = \{A^1_{i,e}, A^2_{i,e}, ...,  A^n_{i,e}\}$.  If we flatten the neural activations, we will obtain a two-dimensional matrix of neural activations for multiple inputs. As shown in Figure \ref{methodology}, by repeating this procedure for all available model snapshots, we can build a three-dimensional tensor containing the neural activations across time for several inputs. Along one dimension of the tensor, one would vary the model snapshot $\{i,e\}$, and along another one would vary the input $m$. %To obtain neural activations for this analysis, we must feed the model with a meaningful input. 
To gather neural activations for this analysis, it is essential to feed meaningful input into the model.
Future sections discuss potential strategies for selecting inputs. %However, we note that our analysis is not limited to neural activations.
However, we emphasize that our analysis extends beyond neural activations.

\subsection{Tensor Component Analysis}

TCA approximates a three-way tensor \(X\) as a sum of rank-1 tensors, where each rank-1 tensor is an outer product of vectors. The three-way tensor can be expressed as follows \citep{williams2018unsupervised}:

\begin{equation}
    \hat{X} \approx \sum_{r=1}^{R} (\mathbf{u^r} \otimes \mathbf{v^r} \otimes \mathbf{w^r})
\end{equation}

In the formulation above, the vector \(\mathbf{u^r}\) might represent patterns in neural activations, \(\mathbf{v^r}\) represents the inputs corresponding to this activity, and \(\mathbf{w^r}\) represents the stages of task learning where this activity is present. The rank \(R\), a hyperparameter for this technique, determines the total number of components that approximate the original tensor.

The optimization objective for TCA aims to minimize the reconstruction error between the original tensor \(X\) and its approximation \(\hat{X}\), defined by the Frobenius norm:

\begin{equation}
    \text{minimize} \quad || X - \hat{X} ||_{F}^2
\end{equation}

Additionally, the objective may be subject to non-negativity constraints on the component vectors \(\mathbf{u^r}\), \(\mathbf{v^r}\), and \(\mathbf{w^r}\), i.e., all elements in these vectors are non-negative.

We will select a low rank \(R\) with reasonable reconstruction error, as described in Section \ref{sec:protocol}.

% Although TCA is introduced in the context of neural recordings \citep{williams2018unsupervised}, in the sections below, we also propose to substitute flattened neural activations for flattened optimized images and similarity vectors to aid in exploring other hypotheses. Overall, our experiments can be summarized as using three-dimensional component analysis, with the aim of tracing out temporal patterns in the internal representations of continually learning models.

\subsection{Activation Tracking}
\label{sec:tracking}
% \fixme{Fix the Activation Tracking section to remove reference to random sampling and class optimized image}

\textit{Hypothesis 1: Does tracking neural activations throughout a continual learning setting, which constrains parameter changes based on an importance measure, reveal specialized classes of neurons?}

Importance-based methods for CL, such as Elastic Weight Consolidation (EWC) \citep{kirkpatrick2017overcoming}, measure the importance of each parameter for a particular task and discourage certain parameters from significant changes throughout training or find meaningful masks for specific tasks. %Our experiments ask whether this class of strategies consequently lead to the emergence of sets of neurons that are specialized for certain tasks. 
In our experiments, we investigate whether these strategies ultimately result in the emergence of sets of neurons that exhibit specialization for particular tasks.
To explore this, we build a tensor with dimensions: [flattened neural activations] x [representative input(s) corresponding to dataset classes] x [model snapshot (time)]. We propose to conduct TCA on this tensor, thus expecting to capture patterns in neural activation of all the selected classes in a dataset across time. 

In order to select the representative input(s) per class for which we will capture neural activations, we will experiment with the following approaches: 
% we will employ the following approach:

\noindent \textbf{Random Sampling} We will randomly select 20 images per class from the test dataset as the class representative inputs.

\noindent \textbf{Maximally Activating Example} We will search the test dataset for an image that maximally activates the desired class in the final model snapshot associated with the task that includes the class. For instance, if class 8 is a part of task $\mathcal{T}_t$ trained over $e$ epochs, we would seek the test image that maximally activates the probability of class 8 in the model snapshot $\theta_{t, e}$.

\noindent \textbf{Class Optimized Image} We will optimize an image for a fixed number of iterations, such that the resulting image maximizes the probability of a particular class, as described by \cite{olah2017feature} for convolutional neural networks (CNNs).

%When looking at a TCA component, we hypothesize that “specialization” may present itself as a cluster of neurons that have higher activations, for a distinct segment of the temporal factors and for one to few inputs. This would signify that certain neurons were highly active in a certain phase of learning when the model was fed certain inputs.
When examining a TCA component, we hypothesize that specialization would be identified as a cluster of neurons characterized by greater activations. These activations would be associated with a segment of the temporal factors and one or few distinct inputs. This would indicate that certain neurons exhibited heightened activity during particular learning stages when the model was exposed to specific inputs.

\subsection{Filter Evolution}
\label{sec:filter}
\textit{Hypothesis 2: Do CNN filters or transformer features within the same layer demonstrate equal levels of activity in their evolution throughout the continual learning process?}

%Are CNN filters or transformer features, within the same layer, equally active in how they evolve throughout the continually learning regime?

In this experiment, we wish to understand how individual CNN filters and transformer features for a chosen layer evolve across the training regime. We ask if certain filters/features update earlier in the training regime and others update later or if filters/features are constantly changing. %In particular, do importance-based methods elicit a starkly different updating pattern compared to a purely replay-based approach or a baseline approach? 
Specifically, do importance-based methods induce a markedly distinct updating pattern when compared to a purely replay-based approach or a baseline approach?
Here, we build our tensor using dimensions: [flattened optimized image] x [filters/features] x [model snapshot (time)]. 

%In contrast to the previous tensor formulation, we will not conduct this analysis for any particular class. Instead of looking across several classes, we look across all the filters or features of a chosen layer in the model. 
In contrast to the previous tensor formulation, this analysis will not be conducted for any specific class. Instead of examining multiple classes, we will focus on studying all the filters or features within a selected layer of the model.
%Further, in place of using neural activations, we plan to use the optimized images for each filter (for CNNs) or feature (for transformers) for a given layer. Similar to how one may optimize images to maximally activate a target class, we will optimize images to maximally activate a selected filter or feature in a model layer, as is conducted for CNN architectures by \cite{olah2017feature} and for transformer architectures by \cite{ghiasi2022vitlearn}.
Additionally, rather than relying on neural activations, our approach involves utilizing optimized images specifically designed for each filter (in the case of CNNs) or feature (in the case of transformers) within a given layer. This optimization process is akin to the method of generating images that maximally activate a target class. In our case, we optimize images to achieve maximum activation for a chosen filter or feature within a model layer. This optimization procedure is inspired by the work of \cite{olah2017feature} for CNN architectures and \cite{ghiasi2022vitlearn} for transformer architectures.

%In this experiment, we may observe that for one of the components, a selection of one particular filter in the filter factors, a selection for a certain segment in the temporal factors, and some subset of pixels in the `flattened image` factors, describing how a filter is changing across time. A result such as this would suggest that one of the filters became active after learning a certain task. The temporal factors would show when it became active, the filter factors would show which filter became active, and the flattened image factors would show whether a certain region or the entire image became active.

In this experiment, we may observe a scenario where a specific component exhibits the following characteristics: it selects one particular filter in the filter factors, corresponds to a particular segment in the temporal factors, and encompasses a subset of pixels within the 'flattened image' factors. %This configuration would essentially describe how a particular filter evolves over time. 
Such a result could imply that this specific filter became actively engaged after learning a certain task. The temporal factors would indicate when it became active, the filter factors would specify which filter was active, and the flattened image factors would reveal which region was active during this process.

%% file: v4/4_experiment.tex
\section{EXPERIMENTAL PROTOCOL}
\label{sec:protocol}

% \begin{table*}[!t]
% \caption{CL Strategies Used in our Analyses} \label{tab:strategies}
% \begin{center}
% \begin{tabular}{p{4.3cm}p{3.0cm}p{7cm}p{1cm}}
% \textbf{Strategy}  &\textbf{Potential SplitCIFAR Params}  &\textbf{Note} &\textbf{+ ER?} \\
% \hline \\
% Naive & - &Lower baseline, train only on current task &No \\
% Cumulative & - & Upper baseline, train on all seen tasks &No \\
% ER & 100 samples per class & Used by \cite{van2022three} & No \\
% MAS \citep{aljundi2018memory} &$\lambda = 4$& Used by \cite{jung2020ags, kaushik2021rmn} &Yes \\

% EWC \citep{kirkpatrick2017overcoming} &$\lambda = 10000$ &Used by \cite{jung2020ags, kaushik2021rmn} & Yes \\

% SI \citep{zenke2017synaptic} &$\lambda = 1$ &Used by \cite{jung2020ags, kaushik2021rmn} & Yes \\

% RMN \cite{kaushik2021rmn} &$\lambda = 0.01$ &Used by \cite{kaushik2021rmn} &Yes \\
% WS \citep{kang2022wsn} &$\lambda = 0.3$ &Used by \cite{kang2022wsn} &Yes \\
% \end{tabular}
% \end{center}
% \end{table*}

\paragraph{Datasets} We propose to run our experiments on the following classification datasets: (i) SplitMNIST, (ii) SplitCIFAR10, (iii) SplitCIFAR100, and (iv) twenty CIFAR100 superclasses \citep{ramasesh2020anatomy}. The proposed settings cover a variety of task complexities. 
Across all comparable experiments, the dataset splits and task orders will be consistent to ensure a fair comparison.
% Further, the latter two scenarios, as explained by \cite{ramasesh2020anatomy}, tackle the challenge of a shift in input distribution by maintaining the same superclasses throughout the tasks, but substituting the underlying subclasses between tasks. 

\paragraph {Task Generation and Order} Influenced by the approach of \cite{ramasesh2020anatomy} and \cite{zhang2022feature}, we wish to apply our framework in a controlled setting by curating the order of tasks. Since we are interested in studying how learned features evolve, we propose that the initial task be large enough to learn rich features.  
We hypothesize that smaller initial tasks lead to poor initial model representations, making it difficult to draw meaningful conclusions. For instance, using Split-CIFAR10, the initial task may consist of the following four classes: airplane, automobile, cat, and horse, allowing the model to `generalize.' We can then curate the next task to either be a two-way classification between deer/dog (animal) or truck/ship (vehicle). We can conduct a similar task curation for SplitMNIST and SplitCIFAR. In order to meaningfully curate these tasks, we propose generating tSimCNE embeddings of the datasets, providing us with a coarse understanding of which classes might share features. Then, for instance, we can select dissimilar classes to consist the initial task and experiment with the remaining classes for the next tasks.

\paragraph{Strategies}
% \textcolor{red}{Sorry, we need to save space}\\
To study how various CL strategies affect model weights, this work will focus on the strategies outlined in Table \ref{table:strategies}. The selected strategies all exploit parameter importance to improve incremental learning, likely being an interesting class of strategies to study for our hypotheses. The final column describes whether there will be an additional variant of the strategy where it is combined with experience replay. We expect to finalize precise hyperparameters through a rigorous grid search.
\paragraph{Model Architectures}

We aim to study three architectures, selecting variants with a similar number of parameters: (i) ResNet-50 (23M parameters) \citep{he2016resnet}, (ii) DeiTSmall (22M parameters) \citep{dosovitskiy2020vit}, and (iii) Convolutional vision Transformer CvT13 (20M parameters) \citep{wu2021cvt}. For the experiments outlined in Sections \ref{sec:tracking} and \ref{sec:filter}, we will use multi-head models and provide task identities during the inference phase. We will train all models with the SGD optimizer using a batch size of 256 images and a fixed learning rate, which will be determined for each architecture by tracking the most accurate model on the validation sets of the scenarios. We expect to train models for a varying number of epochs depending on the architecture and/or the dataset (e.g. 40 epochs for Split-CIFAR-10). We will ensure that certain model training parameters (e.g. number of epochs) are constant across strategies when comparing experimental results. We will equivalently seed models to ensure they are initialized with the same weights.

\paragraph{Hyperparameter Selection Experiments}

We plan to conduct extensive grid search experiments for hyperparameter selection, similar to those conducted by \cite{van2022three}, to ensure that the selected hyperparameters for the main experiments result in models with comparable performance across tasks. The primary focus of this work is to explore how internal representations are affected by the choice of continual learning strategy. Therefore, we will minimize explorations on the effects of various hyperparameters, instead aiming to pick a fair set of hyperparameters that will ensure reasonable and comparable model performance for a given architecture and dataset across continual learning strategies. 

\paragraph{TCA Model Optimization}

We will explore fitting our TCA models with the nonnegative hierarchical alternating least squares (HALS) \citep{cichocki2007hierarchical} and nonnegative block coordinate descent (BCD) algorithms \citep{lee2000algorithms}, with the expectation to select the optimization algorithm that obtains the lowest minima solution for each experiment. We will remain consistent on our selection for a given architecture and dataset. We expect the nonnegative variant of TCA to provide the most interpretable results.

\paragraph{TCA Rank Selection}

Adopting from how \cite{mcguire2022visual} utilized TCA, for each experiment, we plan to fit TCA models across a range of ranks (e.g., 1-20) and plot the errors. We will empirically determine a narrower range of ranks where the ``elbow" in the plot lies, essentially looking for the lowest number of ranks that provide reasonable error. For each rank within the determined ``elbow-range", we will fit 10 TCA models and compare the similarities between the components returned. For the final results, we will select the lowest rank that consistently returns a high similarity (above 0.8) \citep{williams2018unsupervised}.

\paragraph{Evaluation and Metrics}

For all experiments, we plan to report the average of the final classification accuracy. We recognize that our framework is skewed towards an empirical analysis, as we expect to extract activation and filter update patterns from plotting the resulting components after a tensor decomposition. Given that our hypotheses seek some level of specialization within the model, we propose to run masking experiments and measure the change in final accuracy to validate our findings from the TCA models.
% Considering that our hypotheses revolve around uncovering indications of specialization within the model, we aim to validate the practical relevance of our findings from the analysis. To assess the usefulness of our results, we intend to conduct masking experiments and quantify the impact on final accuracy. 
To elaborate, we can randomly mask certain filters in a CNN layer and measure the change in output accuracy. %For instance, if our analysis observes that certain filters are important for a particular class, the masking experiment could may quantify the usefulness of our observation.
If our analysis reveals that specific filters hold significance for a particular class, the subsequent masking experiment could quantify the practical usefulness of our observation.

To compare the components from two TCA models, we will utilize the similarity score proposed in \citep{williams2018unsupervised}, which first solves an assignment problem to match components between two models and finds the ‘optimal’ permutation of components. Then, it computes the dot product between the matched components and returns the mean as a similarity score. 

\begin{figure*}[!t]
\centering
\includegraphics[width=\textwidth]{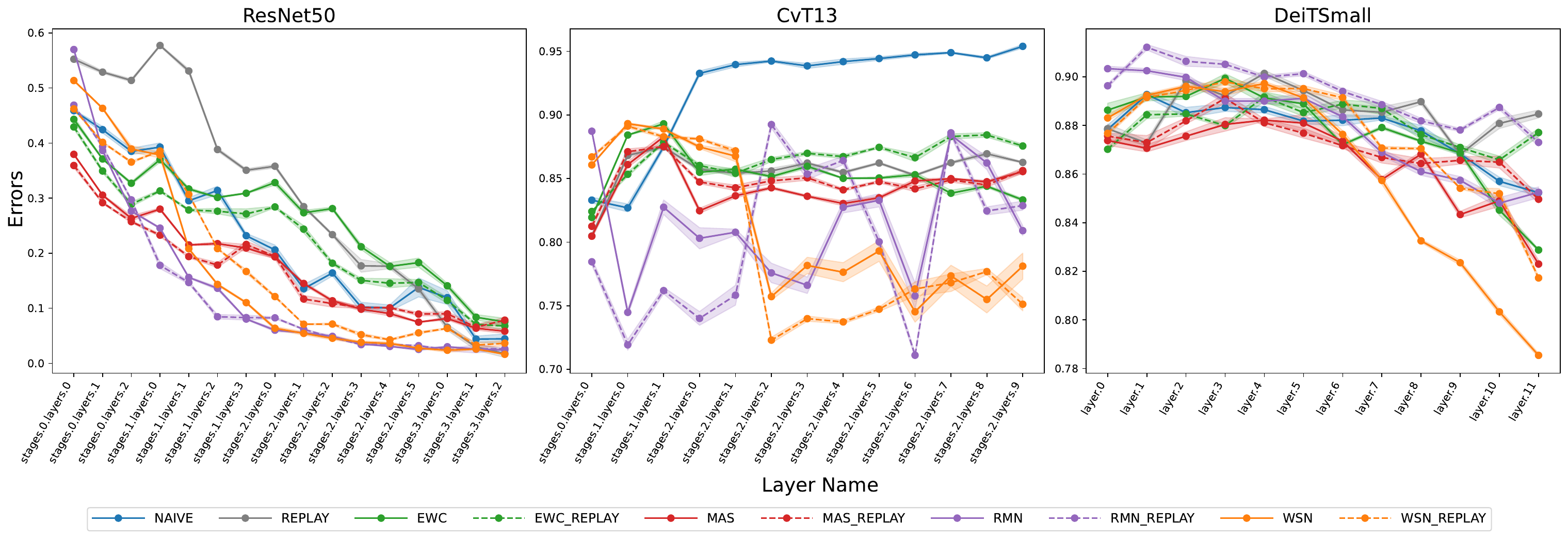}
\caption{Layer-wise reconstruction errors with shaded standard deviations for ResNet50 (\textit{left}), CvT13 (\textit{middle}), and DeiTSmall (\textit{right}) models. \textbf{Note:} The y-axis scaling varies between plots for clearer visualization.} %We should address the y-axis scaling
\label{fig:layer_errors}
\end{figure*}

%% file: v4/5_changes.tex
\section{CHANGES TO THE PROTOCOL}
\label{sec:changes}
% \paragraph{Representative Inputs Selection Technique.} 
% We initially considered three techniques for selecting representative inputs per class (see Section \ref{sec:tracking}): random sampling, maximally activating examples, and class-optimized samples. Ultimately, we opted for maximally activating examples due to computational constraints, as the class-optimized approach proved to be computationally expensive and the random sampling ... \fixme{say the reason}
\paragraph{Strategy Choice.} 
We initially proposed studying the AGS-CL and SI strategies with our method. With regards to the former, we learned that the AGS-CL strategy is not compatible with architectures that include residual streams. To maintain consistency in our approach, we decided to remove this strategy from our study, as it could not be applied to the architectures outlined in Section \ref{sec:protocol}. For the latter, given that we studied the very similar EWC and MAS strategies, we found the inclusion of SI to be redundant and uninformative.
%\fixme{All references to the SI and AGS-CL strategy dataset must be removed.}

\paragraph{Dataset.} While the initial proposal included using the Split-MNIST dataset, we decided against its inclusion, due to the dataset's overly simple nature for our selected architectures. 
% \fixme{All references to SplitMNIST dataset must be removed.}

\paragraph{Representative Input Selection.} Our experiments selected representative inputs by finding maximally activating examples, as outlined in Section \ref{sec:tracking}. Originally, we proposed additional methods for selecting representative inputs, which involved randomly selecting images from a class as well as optimizing noise to maximize the output logits for a particular class, originally described by \cite{olah2017feature}. We did not pursue these methods. We discovered that the latter was an especially computationally intensive procedure and determined that it was not the focus of our study.
% \fixme{Fix the Activation Tracking section to remove reference to random sampling and class optimized image}

%% file: v4/6_results.tex
\begin{figure*}[!t]
\centering
\includegraphics[width=0.48\textwidth]{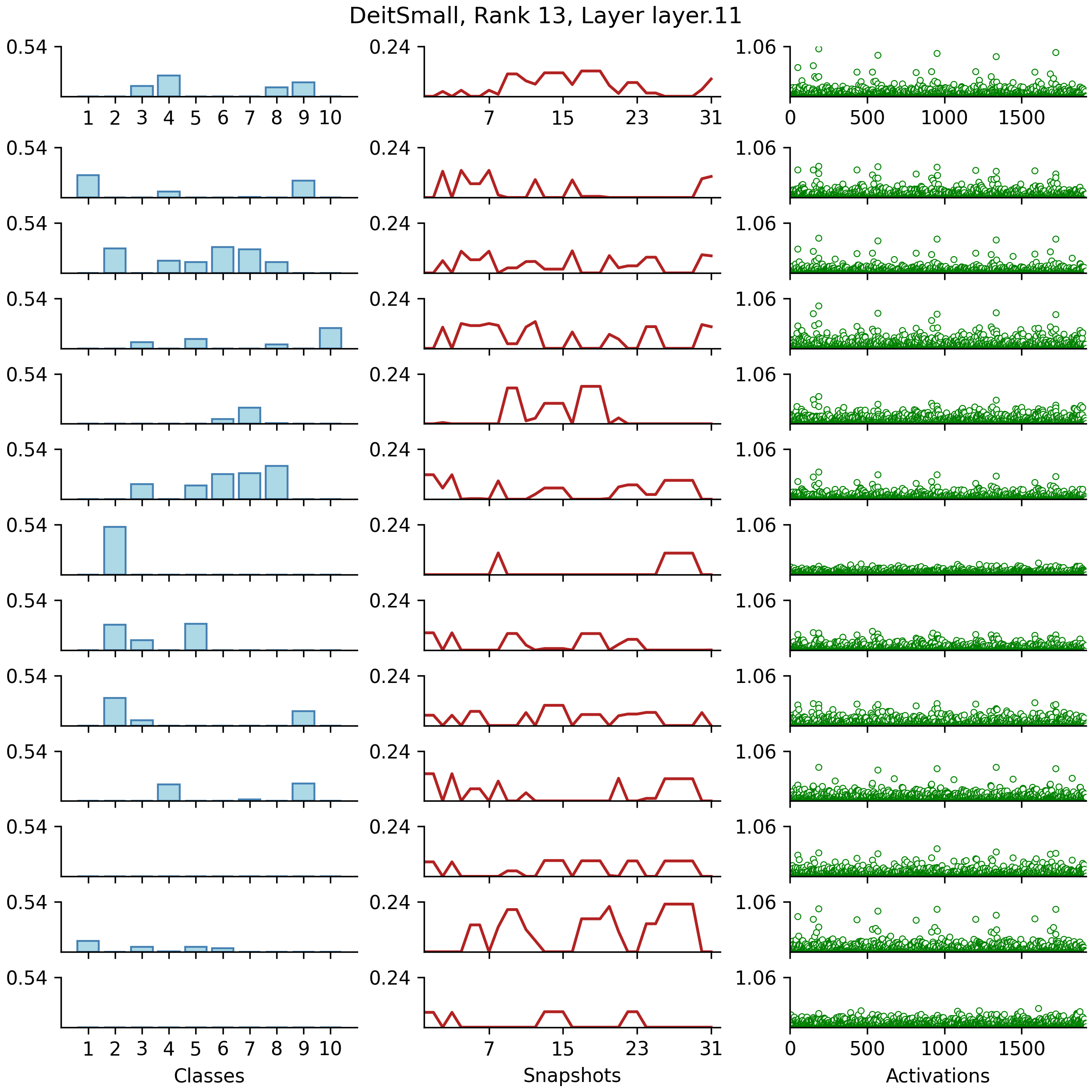}
\includegraphics[width=0.48\textwidth]{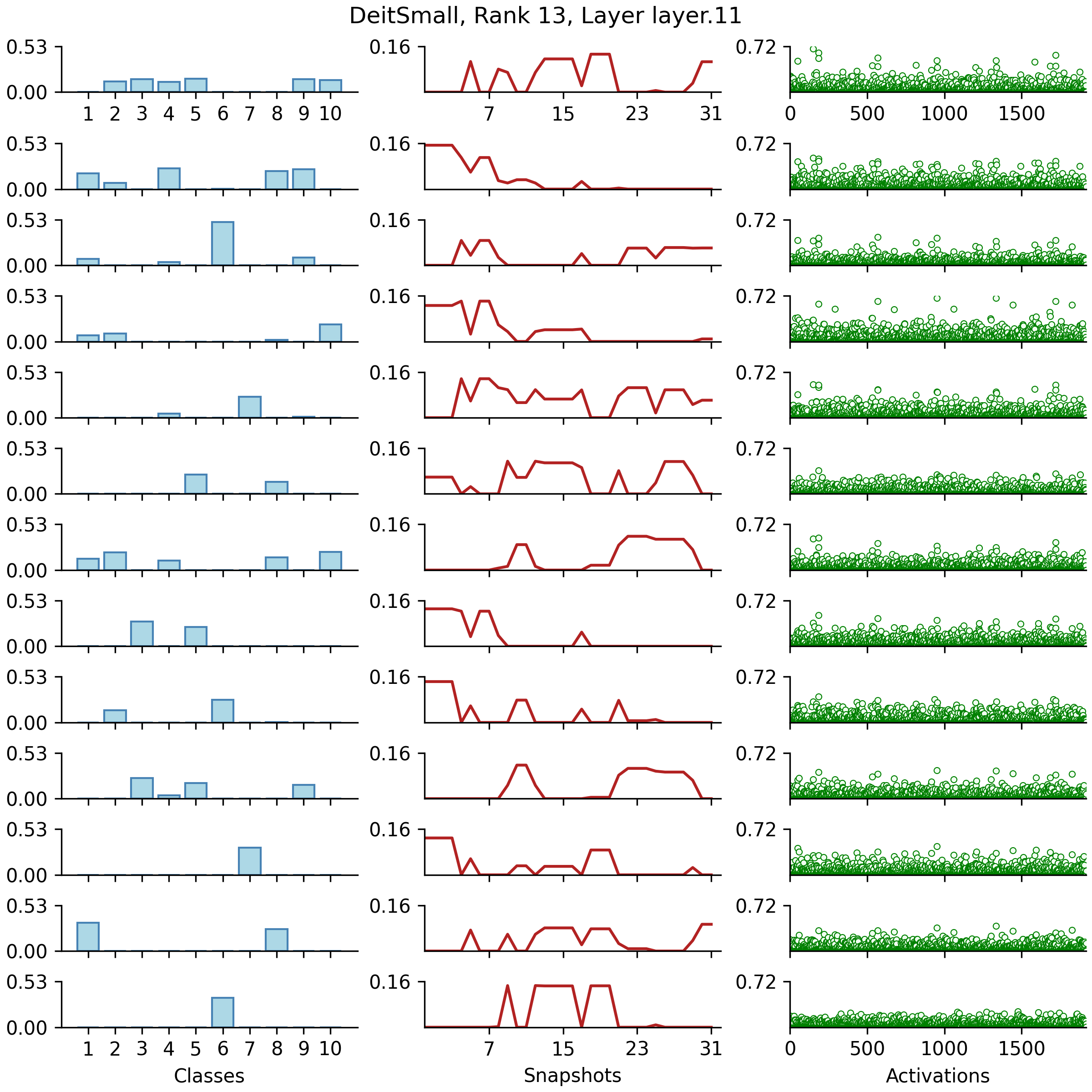}
\caption{Example of tensor component analysis on an activations tensor: two rank-13 TCA plots of activation tensors from DeitSmall trained on Split-CIFAR-10, permuted to order factors that are best aligned. The similarity score between the two is 0.58. \textit{Left}-TCA plot of activations from the MAS strategy, with a reconstruction error of 0.82. \textit{Right}-TCA plot of activation from the MAS Replay strategy, with a reconstruction error of 0.85. Despite the similar performance of the TCA models, as demonstrated in Figure \ref{fig:layer_errors}, the TCA models do not have a high similarity.}
\label{fig:sample_tca_plot_main}
\end{figure*}

\begin{figure*}[!t]
\centering
\includegraphics[width=0.48\textwidth]{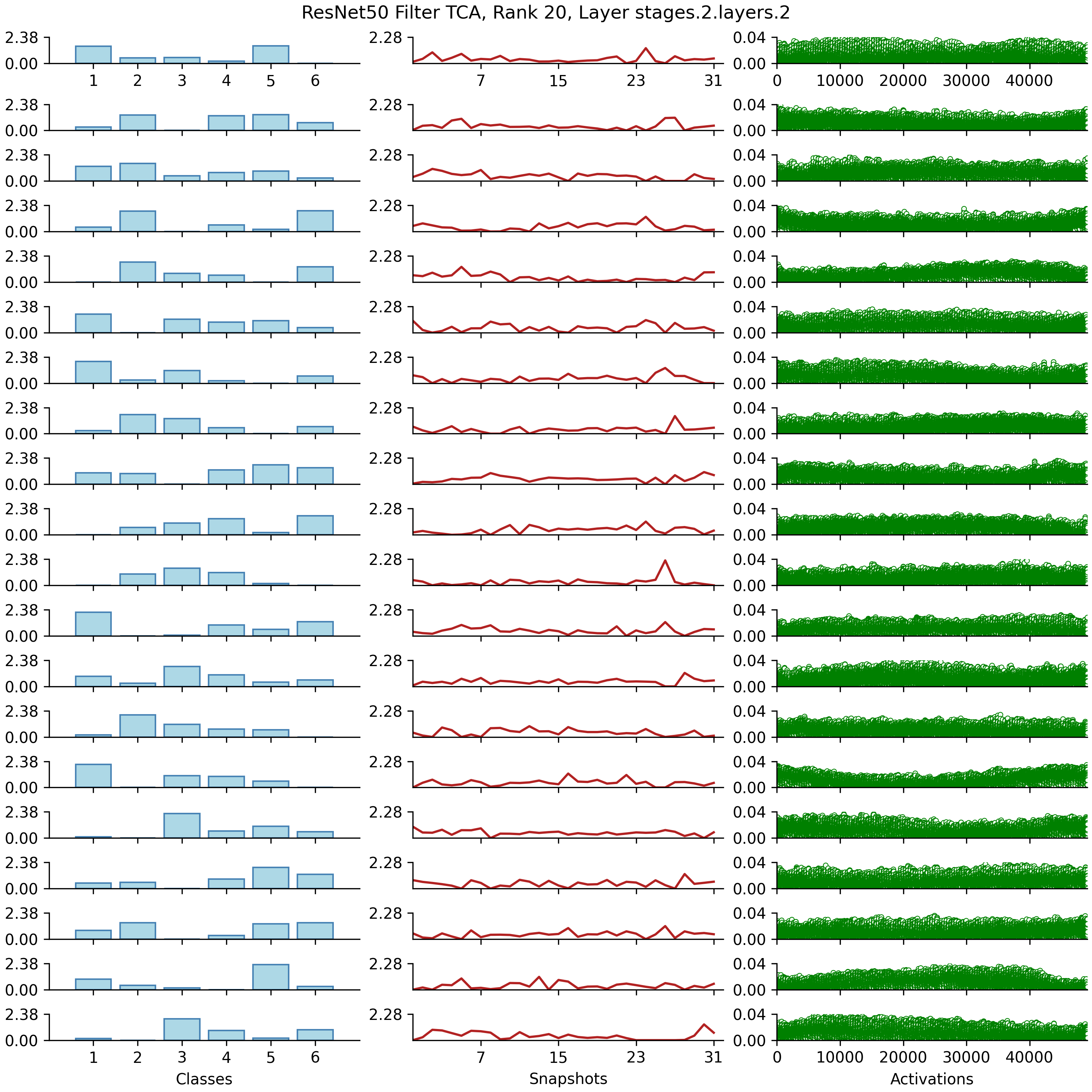}
\includegraphics[width=0.48\textwidth]{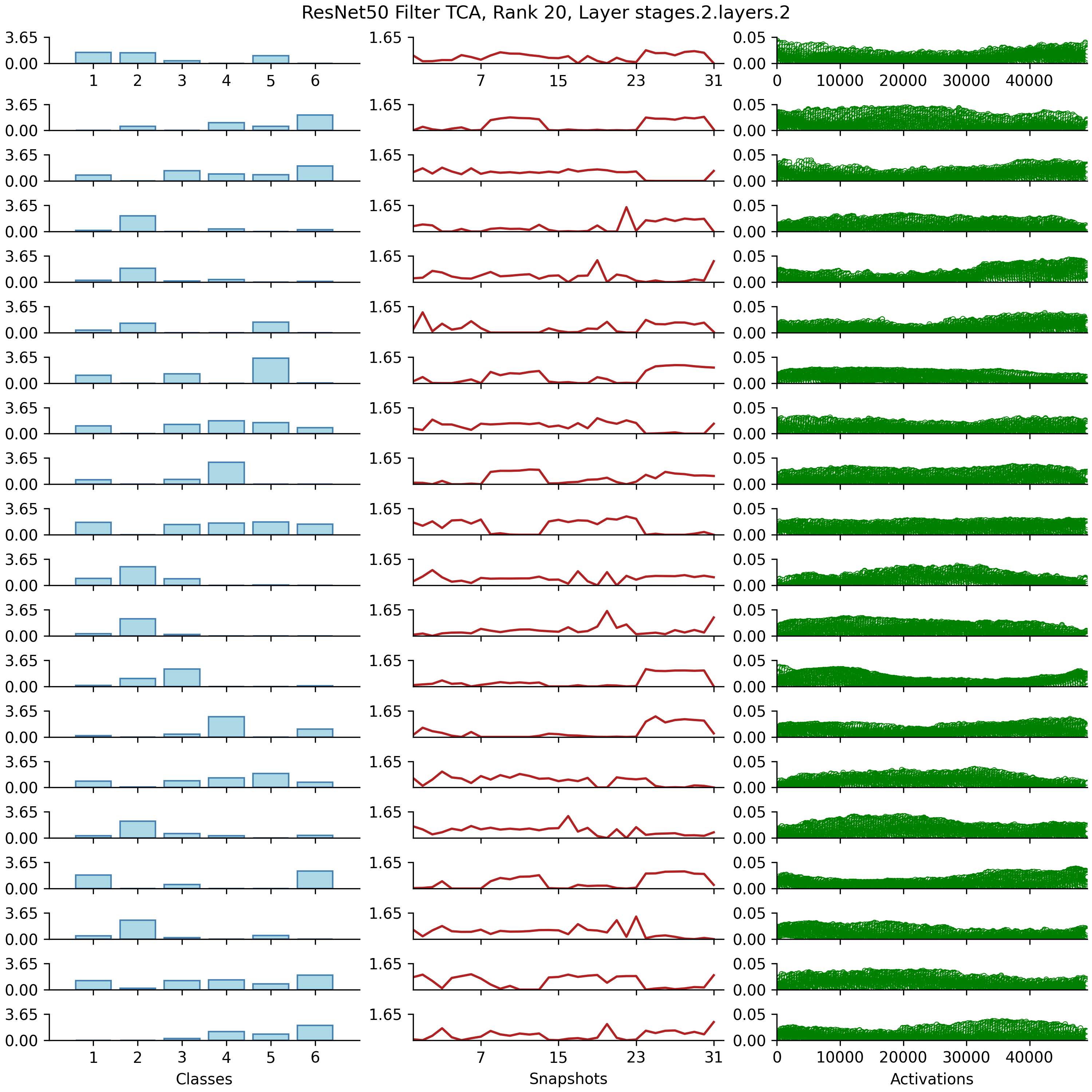}
\caption{Example of tensor component analysis on an optimized images for filters tensor: two rank-20 TCA plots of optimized images for filters tensors from ResNet50 trained on Split-CIFAR-10. \textit{Left}-TCA plot of filter images from the Naive strategy, with a reconstruction error of 0.24. \textit{Right}-TCA plot of activation from the WSN strategy, with a reconstruction error of 0.19.}
\label{fig:sample_filter_tca_plot_main}
\end{figure*}

\section{RESULTS AND DISCUSSIONS}
\label{sec:results}
In this section, we present the results of our experiments along with analyses and discussion.

\subsection{Experimental Setup}

\paragraph{Hardware Configuration.} Our models were trained on a machine equipped with NVIDIA RTX A6000 GPUs, featuring 48GB of GPU memory. 

\paragraph{Tasks Generation.} 
To curate the first task (i.e., Task 1) for each dataset, we used tSimCNE \citep{boehm2023unsupervised} embeddings in conjunction with the QuickHull \citep{barber1996quickhull} algorithm to select a specific number of classes. A detailed description of this process is provided in Appendix \ref{subsec:task_generation}. For the Split-CIFAR-10 benchmark, we selected 4 classes: automobile, airplane, frog, and dog. For Split-CIFAR-100, we selected 10 classes: sea, cloud, aquarium fish, sunflower, lion, raccoon, girl, pickup truck, wardrobe, and chair. For Split-CIFAR-100 superclasses, we selected 4 classes: large outdoor natural scenes, large carnivores, people, and household furniture. The remaining classes were evenly and randomly distributed among the remaining tasks.

\paragraph{Model Training and Hyperparameter Optimization.} Training sessions extended to 120 epochs for both the Split-CIFAR-100 and Split-CIFAR-100-Super datasets, while Split-CIFAR-10 training was for 40 epochs. Hyperparameters were determined through random sweeps across two tasks, using SGD with a momentum of 0.9. Further details on this process can be found in Appendix \ref{sec:suppl_sweeping}. Sweeps were conducted on epochs, batch size, learning rate for different strategies and across different datasets. Table \ref{table:final_hyperparams} provides a breakdown of the final hyperparameters selected for each model and dataset combination.

\subsection{TCA reconstruction errors from activations are significantly varied between architectures.}
To develop an understanding of which layers are useful to study across architectures, we computed the reconstruction errors across all the layers for each architecture. Figure \ref{fig:layer_errors} plots the errors for each of the strategies using activations from Split-CIFAR-10. Surprisingly, we see that the reconstruction error for the ResNet50 architecture is significantly lower than the other architectures. And, there is a clear pattern where later layers in the model tend to have lower reconstruction error. The trend of reconstruction errors across the layers of CvT13, unlike ResNet50 and DeiTSmall, seem to be strategy dependent. There are noticeable variations in reconstruction errors across layers for the subnetworking strategies (WSN and RMN), whereas the reconstruction errors remain largely stable for other strategies. Overall, there is a trend of reconstruction errors following strategy performance (see Figure \ref{fig:cifar10_training}); the subnetworking strategies have lower reconstruction errors than the regularization strategeies. In addition, following our methodology, Figure \ref{fig:rank_errors} plots the reconstruction errors of decompositions from activations of two layers, across ranks 10-24 for all the models. 

\subsection{H1: Activation tracking does not clearly highlight specialized classes of neurons.}

As outlined in Section \ref{sec:tracking}, we conduct TCA on activation tensors. We found that our approach to activation tracking produces factors that are difficult to interpret across different architectures, and the interpretability of these factors does not correlate with the performance of the strategy. Our method does not readily identify specialized clusters of neurons across architectures. Initially, we hypothesized that TCA would cleanly select an input, highlight a cluster of activations, and identify a relevant region on the snapshot (temporal) component. Figure \ref{fig:sample_tca_plot_main} shows a sample result from applying our TCA methodology to activation tensors for a selected layer in DeiTSmall, trained on Split-CIFAR-10. Despite both strategies performing very well with the DeiTSmall model, and having similar reconstruction errors, the decompositions are not similar, scoring only 0.58 on the similarity metric proposed by\cite{williams2018unsupervised}. The authors of that work suggested that, for their tasks, similarity values of at least 0.6 indicate empirically similar factors.

Nevertheless, our decompositions on activation tensors are able to isolate individual classes and regions on the temporal axis corresponding to specific tasks. This is demonstrated across all our activation decomposition plots (see Figures \ref{fig:sample_tca_plot_main},\ref{fig:tca_naive_replay_cifar10}, \ref{fig:tca_mas_wsn_cifar100}). For example, in Figure \ref{fig:sample_tca_plot_main}, the third component on the right plot highlights snapshots from the initial and final task and singles out a class on the inputs axis. In the Split-CIFAR-10 task, class 6 falls in the initial task, but the component highlights snapshots in the initial and the final task. Although it might be tempting to interpret such a component as indicating something about the model’s learning dynamics, the decompositions tend to include conflicting and difficult to interpret selections. For instance, the first and most aligned component in both plots selects several classes and snapshots, along with a few strong activations. We believe it would be naive to attempt and directly assign meaning to such a component. While activation TCA plots for DeiTSmall tend to strongly select for certain activations, this quality is not present in the other activation decomposition plots. On the other hand, activation TCA plots for the other architectures seem to better select for individual classes (see Figure \ref{fig:tca_mas_wsn_cifar100}) or snapshots on the temporal axis that belong to the same task (see Figure \ref{fig:tca_naive_replay_cifar10}). Appendix \ref{sec:more_act_tca} features two other activation tensor decomposition with different strategies for CvT13 and ResNet50.

\subsection{H2: Filter decompositions produce smoother and more pronounced outputs, but remain difficult to interpret.}
% Do CNN filters or transformer features
% within the same layer demonstrate equal levels of activity in
% their evolution throughout the continual learning process
Following our tensor construction from Section \ref{sec:filter}, we also produce TCA plots to study the filters. Figure \ref{fig:sample_filter_tca_plot_main} shows two filter decomposition plots for channels $=[0,100,300,500,700,1000]$. The left decomposition is for the Naive strategy, while the right decomposition is for the WSN strategy. From an empirical analysis, we see that the decomposition on the tensor related to WSN has more salient selections, especially in the temporal axis. Comparing the decomposition for the Naive strategy to figures \ref{fig:tca_filter_ewc_mas} and \ref{fig:tca_filter_replay_rmn}, we also see that the filter selections in the CL strategies, aside from Replay, tend to be more sparse (leftmost columns; blue), while there is a more noticeable selection in the image pixels space (rightmost columns; green). Unlike the activation tensor decompositions, the filter decomposition plots 
produce smoother curves on the temporal axis. On the other hand, it is difficult to try and intuit what a selected filter might represent; whereas, in the activation decompositions, there are clearer meanings to be assigned when a class is selected for a certain region on the temporal axis. 

Our methodology does not immediately clarify whether these filters demonstrate similar activity in their evolution. However, the selection of multiple filters for a certain region of pixels might suggest that these filters are working together. Ultimately, while these plots are still difficult to interpret, we believe that the empirical differences in the plots between the Naive/Replay and the other CL strategies suggests that this approach is a more promising avenue than the previous to study internal representations.

Overall, we found this to be a computationally challenging approach, as it required optimizing images for filters in a chosen layer across all snapshots. Hence, we limited these plots to certain filters and for the CIFAR- dataset. In addition, we limited these plots to the ResNet50 architecture, as we found it challenging to adapt the "channel" optimization procedure for the other architectures in a meaningful way. Appendix \ref{sec:more_filter_tca} features two other activation tensor decomposition with different strategies.

%% file: v4/9_future_work.tex
\section{FUTURE WORK}
Despite the unclear nature of our results, we believe this approach still holds merit. Tensor decompositions are a widely used tool throughout multiple domains and often help provide interpretable insight to high dimensional data. Hence, we believe that our approach would benefit from refinement. In particular, we believe adopting a mechanistic interpretability approach would be more beneficial in understanding the failure cases of using TCA to study continual learning dynamics. To elaborate, we expect that conducting decompositions on activations and filter tensors from significantly smaller toy models and controlled toy datasets would highlight the benefits and pitfalls of this approach, providing a better understanding of how to interpret the outputs of a tensor decomposition. 

%% file: v4/7_conclusion.tex
\section{CONCLUSION}

%We have deliberately chosen to work with importance-based strategies because these methods encourage specialization, such that certain parameters of the model are more important for certain tasks. Our analysis is interested in determining whether we are able to track this type of specialization to mechanistically understand if importance-based methods operate in the intended way. 
Our deliberate choice to work with importance-based strategies stems from their explicit goal to encourage specialization, such that certain model parameters establish greater importance for certain tasks. Our analysis is centered on determining whether we are able to track this type of specialization to understand if importance-based methods operate as intended. 
We believe specialization in CL is crucial because it encourages an efficient use of predefined resources as the model learns to allocate parameters to accommodate new tasks. By identifying meaningful behavior of existing methods through this analysis, we wish to pave the way to improve importance-based methods or discover the need to seek a different angle of attack to elicit specialization.
%Our emphasis on specialization in is grounded in the belief that these aspects are pivotal. They promote the efficient utilization of predefined resources by requiring the model to judiciously allocate parameters to adapt to new tasks. By identifying meaningful behaviors of existing methods through this analysis, we pave the way for potential enhancements in importance-based methods or the exploration of alternative approaches to enhance and harness specialization. This research could ultimately contribute to more effective strategies for CL and resource allocation in machine learning models.

Ultimately we found that our methodology did not successfully highlight specialized clusters of neurons or aid in understanding the evolution of filters. Nevertheless, glimpses in our results still demonstrate that this approach has merit and could be refined and better understood for smaller datasets and architectures. By building an intuition from the ground up of what TCA shows us in this context, we believe a variation of our approach could be a beneficial tool for study.

%% file: v4/supplementary_a.tex
% \section*{\centering SUPPLEMENTAL INFORMATION}
% \label{sec:supplementary}

\section{MODEL METHODS}
\label{sec:suppl_model_methods}
%%%%%%%%%%%%%%%%%%%%%%%%%%%%%%%%%%%%%%%%%%%%%%%%%%%%%%%%%%%%%%%%%%%%%%%%%%%%%%%%%%%%%%%
%%%%%%%%%%%%%%%%%%%%%%%%%%%%%%%%%%%%%%%%%%%%%%%%%%%%%%%%%%%%%%%%%%%%%%%%%%%%%%%%%%%%%%%

\subsection{Task Generation and Order}
\label{subsec:task_generation}
\begin{figure*}[!h]
\centering
\includegraphics[width=0.31\textwidth]{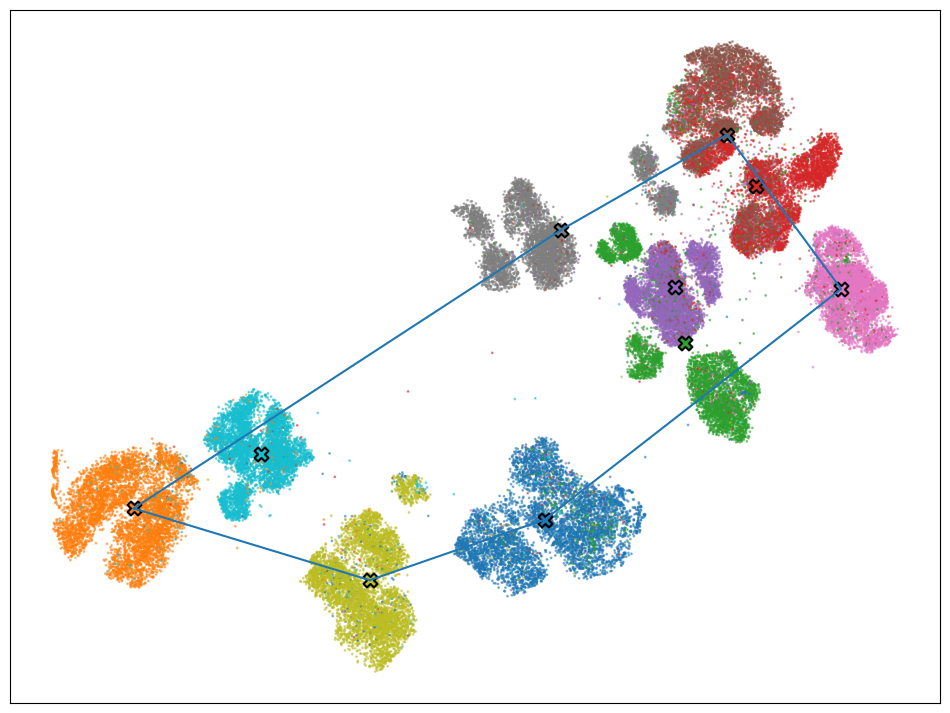}
\hfill 
\includegraphics[width=0.31\textwidth]{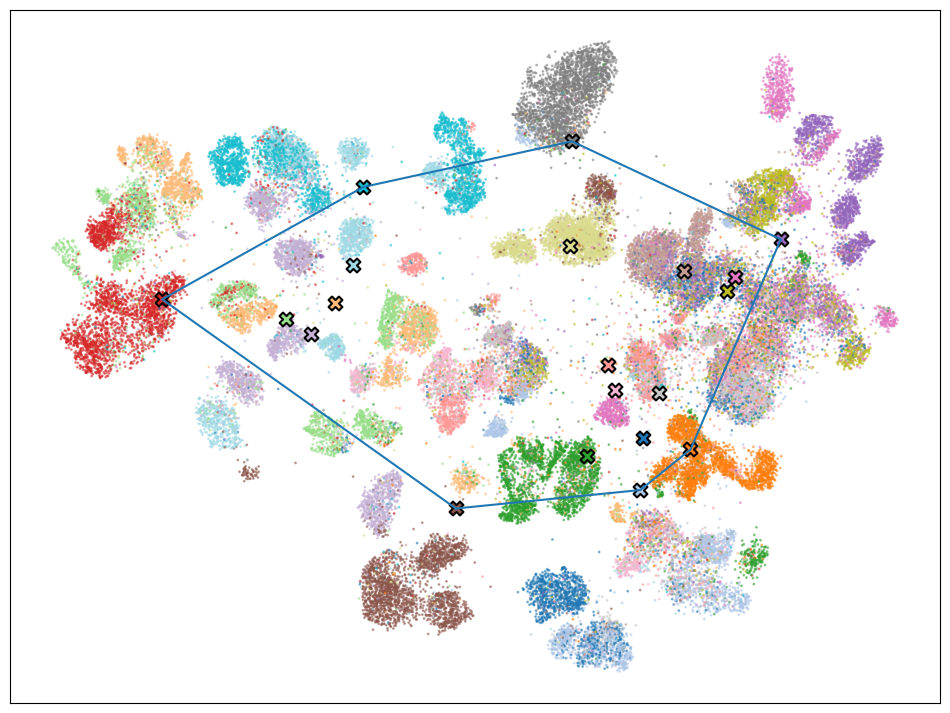}
\hfill
\includegraphics[width=0.31\textwidth]{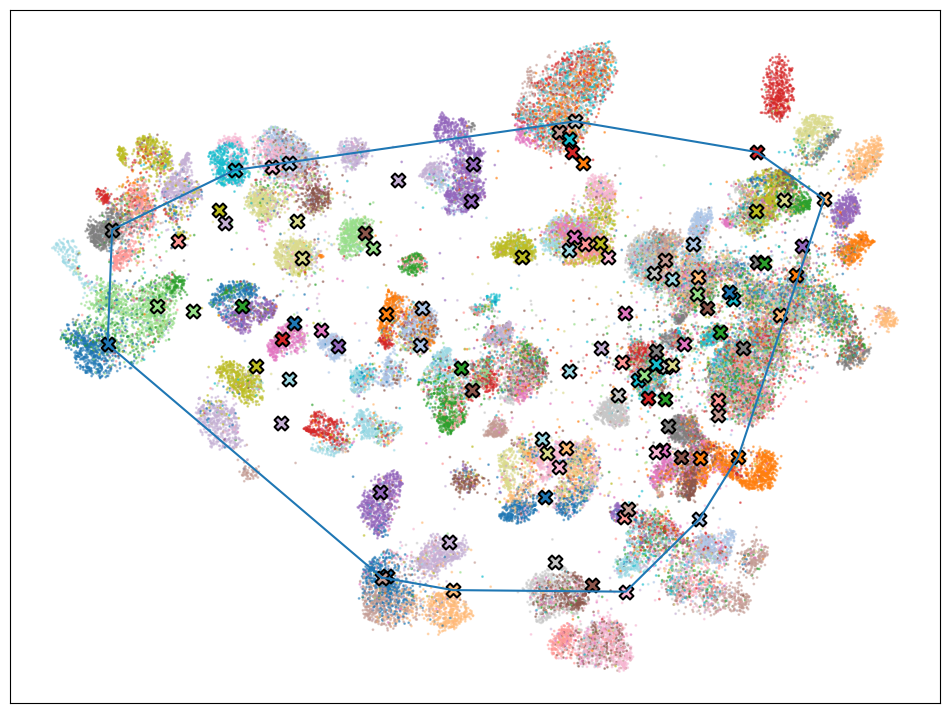}
\caption{Visualization of tSimCNE embeddings, the computed centroids for each class (marked by crosses), and the convex hull constructed using QuickHull algorithm for CIFAR-10 (\textit{left}), CIFAR-100 superclasses (\textit{middle}), and CIFAR-100 (\textit{right}) datasets.}
\label{cifar_embedding}
\end{figure*}

This section elucidates the methodology employed to achieve the curated task generation outlined in Section \ref{sec:protocol}. In summary, we leverage the tSimCNE embeddings of the CIFAR datasets, as introduced by Boehm et al. \citep{boehm2023unsupervised}, in conjunction with the QuickHull algorithm \citep{barber1996quickhull} to delineate the classes for constructing the first task. Figure \ref{cifar_embedding} visualizes the tSimCNE embeddings, along with the computed centroids for each class (denoted by crosses), elucidate the convex hull defined by the subset $M$ for the $N$ centroids.

tSimCNE employs contrastive learning, a technique that enhances an image sample by creating two augmented versions. It then optimizes a CNN to minimize the distance between the latent representations of both augmented versions of the sample. Authors in \citep{boehm2023unsupervised} propose to optimize the following loss function: 
\begin{equation*}
\ell_\text{t-SimCNE}(i,j) = - \log \dfrac{1/(1+\|z_i-z_j\|^2)}{\sum^{2b}_{k\neq i}1/(1+\|z_i-z_k\|^2)}
\end{equation*}

where $z_i, z_j$ refer to two augmentations of the same image, in the latent space. When minimizing the loss, the numerator enforces a similarity between the same sample, while the denominator acts to repel different samples. For CIFAR-10 and CIFAR-100, we use the 2D embeddings obtained by the authors' experiments, made available at \url{https://github.com/berenslab/t-simcne}.

To summarize a class, we first compute the mean $x$ and $y$ coordinates of all samples per class, resulting in a representative centroid for each of the $N$ classes. Following this, we employ the QuickHull algorithm to identify a subset of $M$ centroids, where $M < N$, that constitute the convex hull encompassing all class centroids. For the QuickHull algorithm, we utilize the implementation provided at \url{https://github.com/zifanw/ConvexHull2D}. 

% The QuickHull algorithm identified $M=6$ centroids for CIFAR-10, of which we randomly selected four classes to consist of Task 1. For CIFAR-100 (superclasses), the algorithm selected $M=11$ ($M=7$) centroids, of which we randomly selected ten (four) classes. The remaining classes were evenly split among the remaining tasks. The tasks are listed in Table \ref{table:initial_tasks}.

The QuickHull algorithm identified $M=6$ centroids for CIFAR-10, from which we randomly selected 4 classes to form Task 1. For CIFAR-100 (superclasses), the algorithm identified $M=11$ centroids ($M=7$ for CIFAR-100 superclasses), from which we randomly selected 10 classes (4 for CIFAR-100 superclasses). The remaining classes were evenly distributed among the remaining tasks. The tasks are listed in Table \ref{table:initial_tasks}.

\begin{table}[h!]
\centering
\begin{tabular}{@{}lll@{}} 
\toprule
Dataset & $|\mathcal{T}_1|$ & Classes\\ \midrule
CIFAR-10 & 4 & automobile, airplane, frog, dog\\
CIFAR-100 superclasses & 4 & large outdoor natural scenes, large carnivores, people, household furniture\\
CIFAR-100 & 10 & sea, cloud, aquarium fish, sunflower, lion, raccoon, girl, pickup truck, wardrobe, chair\\ \hline
\end{tabular}
\caption{Number of selected classes for the initial task ($\mathcal{T}_1$) and their corresponding labels across datasets.}
\label{table:initial_tasks}
\end{table}
%%%%%%%%%%%%%%%%%%%%%%%%%%%%%%%%%%%%%%%%%%%%%%%%%%%%%%%%%%%%%%%%%%%%%%%%%%%%%%%%%%%%%%%
%%%%%%%%%%%%%%%%%%%%%%%%%%%%%%%%%%%%%%%%%%%%%%%%%%%%%%%%%%%%%%%%%%%%%%%%%%%%%%%%%%%%%%%

%%%%%%%%%%%%%%%%%%%%%%%%%%%%%%%%%%%%%%%%%%%%%%%%%%%%%%%%%%%%%%%%%%%%%%%%%%%%%%%%%%%%%%%
%%%%%%%%%%%%%%%%%%%%%%%%%%%%%%%%%%%%%%%%%%%%%%%%%%%%%%%%%%%%%%%%%%%%%%%%%%%%%%%%%%%%%%%

\subsection{Hyperparameter Sweeps}
\label{sec:suppl_sweeping}
Our hyperparameter selections were guided by random hyperparameter sweeps for two tasks using SGD with momentum set to 0.9. For each of the three architectures introduced in Section \ref{sec:protocol}, initial sweeps were conducted on epochs, batch size, and learning rate using the Split-CIFAR-10 and Split-CIFAR-100 datasets, employing the Naive strategy. Upon determining the appropriate epochs and batch size for each architecture and dataset combination, additional sweeps were conducted on learning rate and strategy-specific hyperparameters for the contrastive learning (CL) approaches on Split-CIFAR-100. The selected hyperparameters for model training are discussed in Section \ref{sec:suppl_training}.

Tables \ref{table:sweep_basic}, \ref{table:sweep_regs}, and \ref{table:sweep_subnet} detail the sweep configurations and selected hyperparameters. The results of our sweeps are documented at: \url{https://wandb.ai/nishantaswani/cl_decomp/sweeps}. We recommend sorting by the first column for ease of navigation.

% \fixme{(minor) Ideally all scientific notation values should be using the SI package}

\begin{table}[h!]
\centering
\resizebox{\textwidth}{!}{%
\begin{tabular}{@{}lllllll@{}}
\toprule
Strategy & Dataset & Epochs & Batch Size & Learning Rate & Random Samples & Memory Size \\ \midrule
\multirow{2}{*}{Naive} & Split-CIFAR-10 & 30 & LogUniform(64,256) & Uniform(\SI{1e-1}, \SI{5e-2}) & 20 & - \\
 & Split-CIFAR-100 & \{60,90\} & \{64,128,256\} & Uniform(\SI{1e-2}, \SI{4e-2}) & 50 & - \\
Replay & Split-CIFAR-100 & 50 & 90 & LogUniform(\SI{3e-3},\SI{3e-2}) & 50 & QLogUniform(20,500,q=10) \\ \bottomrule
\end{tabular}%
}
\caption{Sweep configurations for the hyperparameters used by the Naive and Replay strategies across various datasets.}
\label{table:sweep_basic}
\end{table}

% \fixme{NOTE: all sweeps for table 4 and 5 (just below me) were ran 50 times, sgd momentum 0.9, 90 epochs, batch size 256. }
The sweeps corresponding to Table \ref{table:sweep_regs} and \ref{table:sweep_subnet} were conducted 50 times each, with a fixed SGD momentum of 0.9, spanning 90 epochs, and employing a batch size of 256. 

% A sweep example for the WSN strategy with DeiT architecture on the Split-CIFAR-100 dataset is presented in Figure \ref{fig:sweep}..

\begin{table}[h!]
\centering
\resizebox{0.7\textwidth}{!}{%
\begin{tabular}{@{}llll@{}}
\toprule
Strategy & Learning Rate & $\lambda$ Parameter & $\alpha$ Parameter \\ \midrule
EWC & LogUniform(\SI{1e-3},\SI{3e-2}) & QLogUniform(1,\SI{1e2},q=1) & - \\
MAS & LogUniform(\SI{1e-3},\SI{3e-2}) & QLogUniform(1,\SI{1e2},q=1) & Uniform(\SI{1e-1}, 1) \\ \bottomrule
\end{tabular}}
\caption{Sweep configurations for the hyperparameters used by the EWC and MAS strategies on Split-CIFAR-100 dataset.}
\label{table:sweep_regs}
\end{table}

\begin{table}[h!]
\centering
\resizebox{\textwidth}{!}{%
\begin{tabular}{@{}lllll@{}}
\toprule
Strategy & Learning Rate & Capacity $\text{param\_c}$ & Prune Epoch & Pruning Threshold (wt\_param) \\ \midrule
WSN & LogUniform(\SI{1e-3},\SI{3e-2}) & QUniform(\SI{1e-1}, \SI{1e0}, q=0.1) & - & - \\
RMN & LogUniform(\SI{1e-3},\SI{3e-2}) & - & QUniform(30,80,q=5) & LogUniform(\SI{1e-2},\SI{1e-1}) \\ \bottomrule
\end{tabular}%
}
\caption{Sweep configurations for the hyperparameters used by the WSN and RMN strategies on Split-CIFAR-100 dataset.}
\label{table:sweep_subnet}
\end{table}

% \begin{figure}[!htp]
% %\vspace{.3in}
% %\centerline{\fbox{This figure intentionally left non-blank}} 
% \centering
% \includegraphics[width=0.65\textwidth, height=7cm]{v4/figures/sweep_example.png}
% %\vspace{.3in}
% \caption{Sweep example for WSN strategy with DeiT architecture on Split-CIFAR-100 dataset. }
% \label{fig:sweep}
% \end{figure}
%%%%%%%%%%%%%%%%%%%%%%%%%%%%%%%%%%%%%%%%%%%%%%%%%%%%%%%%%%%%%%%%%%%%%%%%%%%%%%%%%%%%%%%
%%%%%%%%%%%%%%%%%%%%%%%%%%%%%%%%%%%%%%%%%%%%%%%%%%%%%%%%%%%%%%%%%%%%%%%%%%%%%%%%%%%%%%%

%%%%%%%%%%%%%%%%%%%%%%%%%%%%%%%%%%%%%%%%%%%%%%%%%%%%%%%%%%%%%%%%%%%%%%%%%%%%%%%%%%%%%%%
%%%%%%%%%%%%%%%%%%%%%%%%%%%%%%%%%%%%%%%%%%%%%%%%%%%%%%%%%%%%%%%%%%%%%%%%%%%%%%%%%%%%%%%
\subsection{Model Training}
\label{sec:suppl_training}

% All models were trained on Split-CIFAR-100 and Split-CIFAR-100-Super for 120 epochs. For Split-CIFAR-10, training lasted 40 epochs. Table \ref{table:final_hyperparams} details the final hyperparameters used for each model and dataset combination.

\begin{table}[t!]
\centering
\resizebox{\textwidth}{!}{

\begin{tabular}{@{}lllllll@{}}
\toprule
Strategy & Architecture & Epochs & Batch Size & Learning Rate & Memory Size & Other Hyperparameters \\ \midrule

\multirow{3}{*}{Naive} & CvT13 &  &  & \SI{2.50E-02}{} & \multirow{3}{*}{-} & \multirow{3}{*}{-} \\
& DeiTSmall &  &  & \SI{2.00E-02}{} &  &  \\
& ResNet50 &  &  & \SI{1.00E-02}{} &  &  \\ 
\cmidrule{1-2} \cmidrule{5-7} 

\multirow{3}{*}{Cumulative} & CvT13 &  \multirow{3}{*}{\begin{tabular}[c]{@{}l@{}}$40$ for SC10; \\ $120$ for SC100 \\ and SC100-Super\end{tabular}} & \multirow{3}{*}{256} & \SI{2.50E-02}{} & \multirow{3}{*}{-} & \multirow{3}{*}{-} \\
 & DeiTSmall &  &  & \SI{2.00E-02}{} &  &  \\
 & ResNet50 &  &  & \SI{1.00E-02}{} &  &  \\ 
\cmidrule{1-2} \cmidrule{5-7} 

\multirow{3}{*}{Replay} & CvT13 &  &  & \SI{2.00E-02}{} & \multirow{3}{*}{200}  & \multirow{3}{*}{-} \\
 & DeiTSmall &  &  & \SI{2.00E-02}{} &  &  \\
 & ResNet50 &  &  & \SI{1.50E-02}{} &  &  \\ 
 \cmidrule{1-2} \cmidrule{5-7} 

\multirow{3}{*}{EWC {[}Replay{]}} & CvT13 &  &  & \SI{2.00E-02}{} & \multirow{3}{*}{{[}200{]}} & \multirow{3}{*}{$\lambda = 30$} \\
 & DeiTSmall &  &  & \SI{1.50E-02}{} &  &  \\
 & ResNet50 &  &  & \SI{5.00E-03}{} &  &  \\ 
\cmidrule{1-2} \cmidrule{5-7} 

\multirow{3}{*}{MAS {[}Replay{]}} & CvT13 &  &  & \SI{1.50E-02}{} & \multirow{3}{*}{{[}200{]}} & \multirow{3}{*}{$\lambda = 3, \alpha=0.5$} \\
 & DeiTSmall &  &  & \SI{1.50E-02}{} &  &  \\
 & ResNet50 &  &  & \SI{5.00E-03}{} &  &  \\ 
\cmidrule{1-2} \cmidrule{5-7} 

\multirow{3}{*}{RMN {[}Replay{]}} & CvT13 &  &  & \SI{2.50E-02}{} & \multirow{3}{*}{{[}200{]}} & \multirow{3}{*}{\begin{tabular}{@{}l@{}}
    prune epoch = 15 for SC10; \\
    50 for SC100 and SC100-Super; \\
    wt\_param = \SI{1.00E-2}{} \\
\end{tabular}}
\\
 & DeiTSmall &  &  & \SI{2.00E-02}{} &  &  \\
 & ResNet50 &  &  & \SI{3.00E-03}{} &  &  \\ 

\cmidrule{1-2}  \cmidrule{5-7}

\multirow{3}{*}{WSN {[}Replay{]}} & CvT13 &  &  & \SI{2.00E-02}{} & \multirow{3}{*}{{[}200{]}} & \multirow{3}{*}{capacity (param\_c) $=0.5$} \\
 & DeiTSmall &  &  & \SI{2.00E-02}{} &  &  \\
 & ResNet50 &  &  & \SI{1.00E-02}{} &  &  \\ 
\bottomrule

\end{tabular}
}
\caption{List of final hyperparameters used for training the different architectures on the selected strategies. SC10 refers to the Split-CIFAR-10 dataset. SC100 and SC100-Super refer to the Split-CIFAR-100 and Split-CIFAR-100-Superclasses datasets, respectively.}
\label{table:final_hyperparams}

\end{table}

We used existing implementations from the Avalanche framework \citep{avalanche} for EWC, MAS, and Replay. Due to our use of architectures not originally outlined in the RMN and WSN papers, we had to develop flexible implementations of RMN and WSN. We developed these implementations guided by details from the original papers and publicly available code repositories provided by the respective authors. We will contribute our Avalanche-based implementations of RMN and WSN to the Avalanche community. Table \ref{table:final_hyperparams} details the final hyperparameters used for each model and dataset combination.

%%%%%%%%%%%%%%%%%%%%%%%%%%%%%%%%%%%%%%%%%%%%%%%%%%%%%%%%%%%%%%%%%%%%%%%%%%%%%%%%%%%%%%%
%%%%%%%%%%%%%%%%%%%%%%%%%%%%%%%%%%%%%%%%%%%%%%%%%%%%%%%%%%%%%%%%%%%%%%%%%%%%%%%%%%%%%%%

%% file: v4/supplementary_b.tex
\section{TENSOR COMPONENT ANALYSIS (TCA)}
\label{sec:suppl_tca}
We fit the tensor component analysis (TCA) with the \texttt{tensortools} package \citep{williams2018unsupervised} in Python. From empirical results, we found minimal difference between the nonnegative HALS and nonnegative BCD algorithms for fitting our TCA models. Hence, we selected the nonnegative BCD algorithm to fit our models. 

\subsection{Reconstruction Error per Layer}
We fit $\text{rank}=15$ TCA models across all the layers of each model, for each strategy. Each reconstruction error was averaged across 5 TCA models. We plot the means (circle) and their standard deviations (shading) in Figure \ref{fig:layer_errors}.

% \begin{figure*}[!htbp]
% \centering
% \includegraphics[width=\textwidth]{v4/figures/layer_error.pdf}
% \caption{Layer-wise Reconstruction Errors with Shaded Standard Deviations for ResNet50, CvT13, and DeiTSmall Models. \textbf{Note:} The y-axis scaling varies between plots for better visualization.} %We should address the y-axis scaling
% \label{fig:layer_errors}
% \end{figure*}

\subsection{Rank Search}
\label{subsection_rank_search}
\begin{figure*}[!htbp]
\centering
\includegraphics[width=\textwidth]{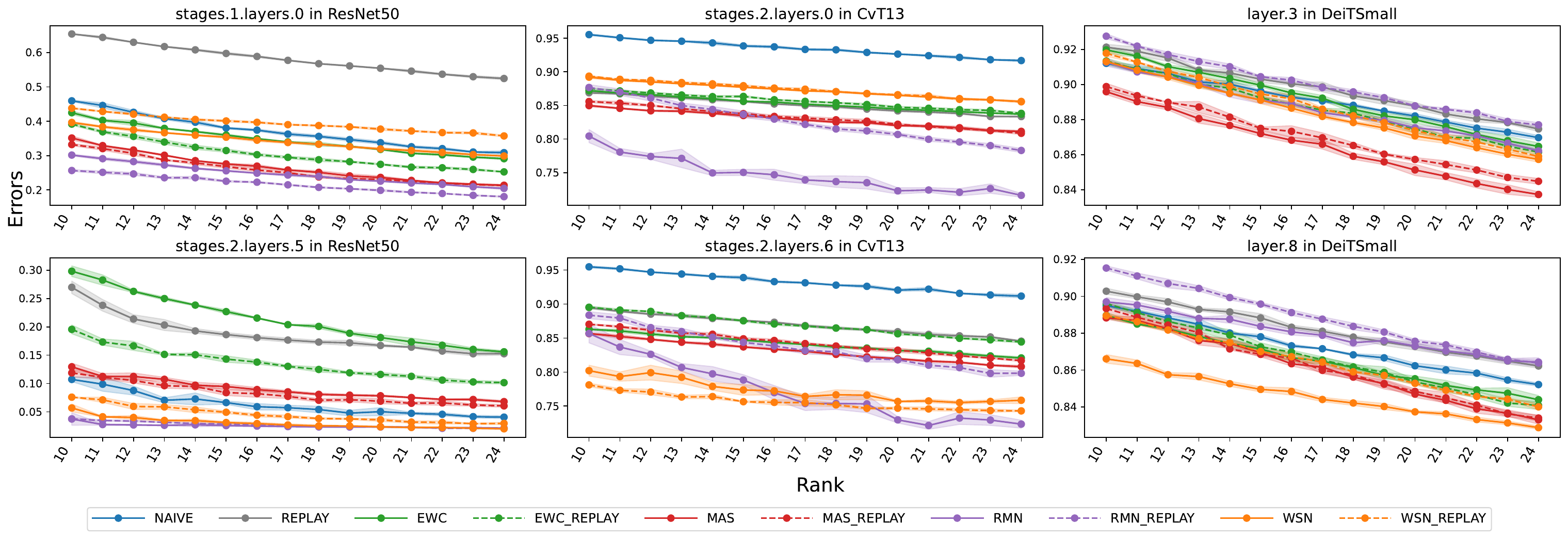}
\caption{Reconstruction errors across ranks with standard deviations for ResNet50 (\textit{left}), CvT13 (\textit{middle}), and DeiTSmall (\textit{right}) models studied on two layers from their respective models. In the top row, we selected an early layer and in the bottom row, a relatively later layer in the architecture. \textbf{Note:} The y-axis scaling varies between plots for clearer visualization.} %We should address the y-axis scaling
\label{fig:rank_errors}
% \locallabel{fig:rank_errors}
\end{figure*}
As outlined in Section \ref{sec:protocol}, we fit TCA models across ranks 10-24 to study the effect of rank selection on the performance of our TCA models.
Figure \ref{fig:rank_errors} shows our resultson CIFAR-10, for two layers per model, based on tensors created from layer activations. Similar to the findings from Figure \ref{fig:layer_errors}, the TCA models are a lot more performant in the case of ResNet50. Aside from EWC and Replay on ResNet50, and RMN on CvT13, the TCA models do not improve drastically when increasing the rank of the models.

\subsection{Additional Activation Tensor Decomposition Plots}
\label{sec:more_act_tca}
\begin{figure*}[!htbp]
\centering
\includegraphics[width=0.48\textwidth]{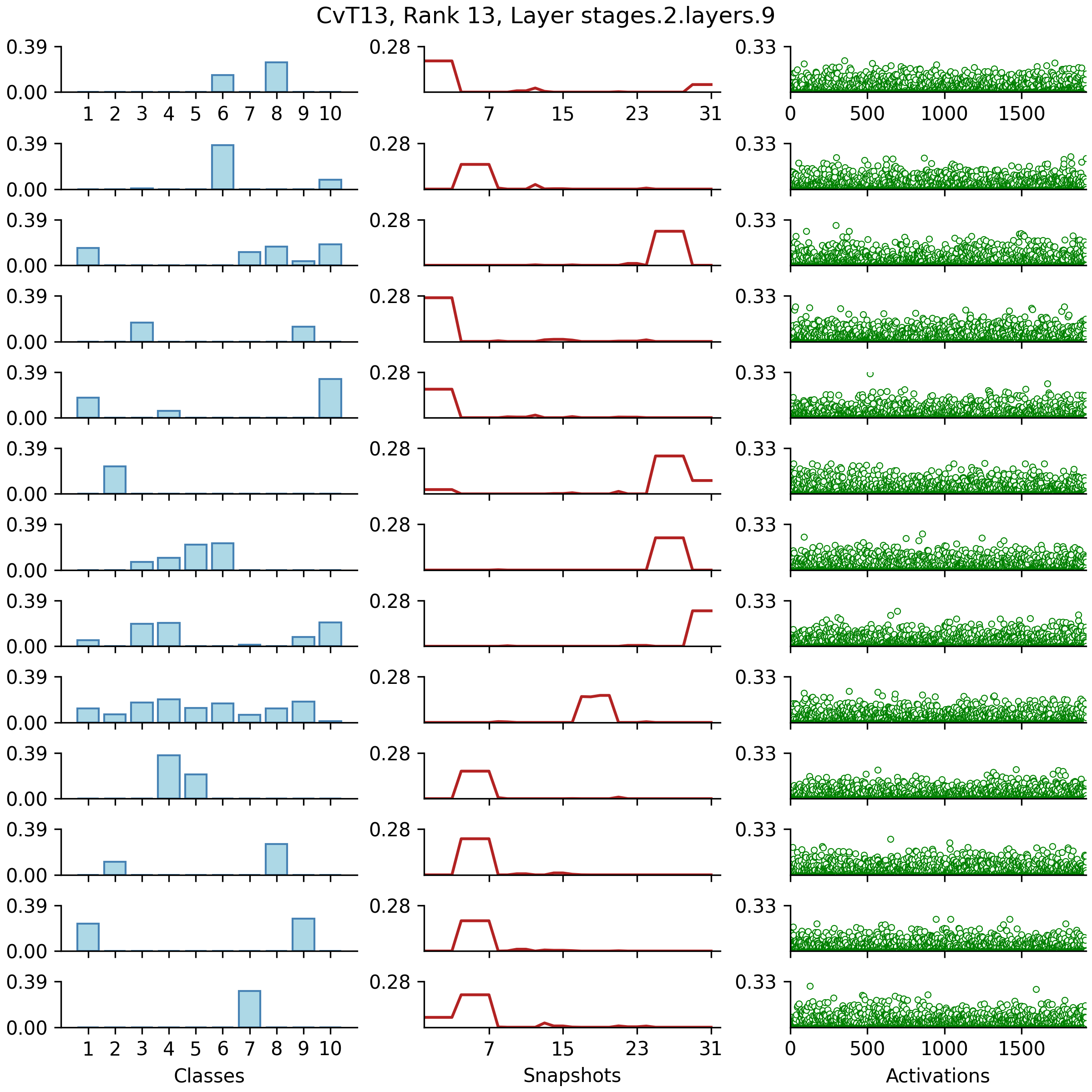}
\includegraphics[width=0.48\textwidth]{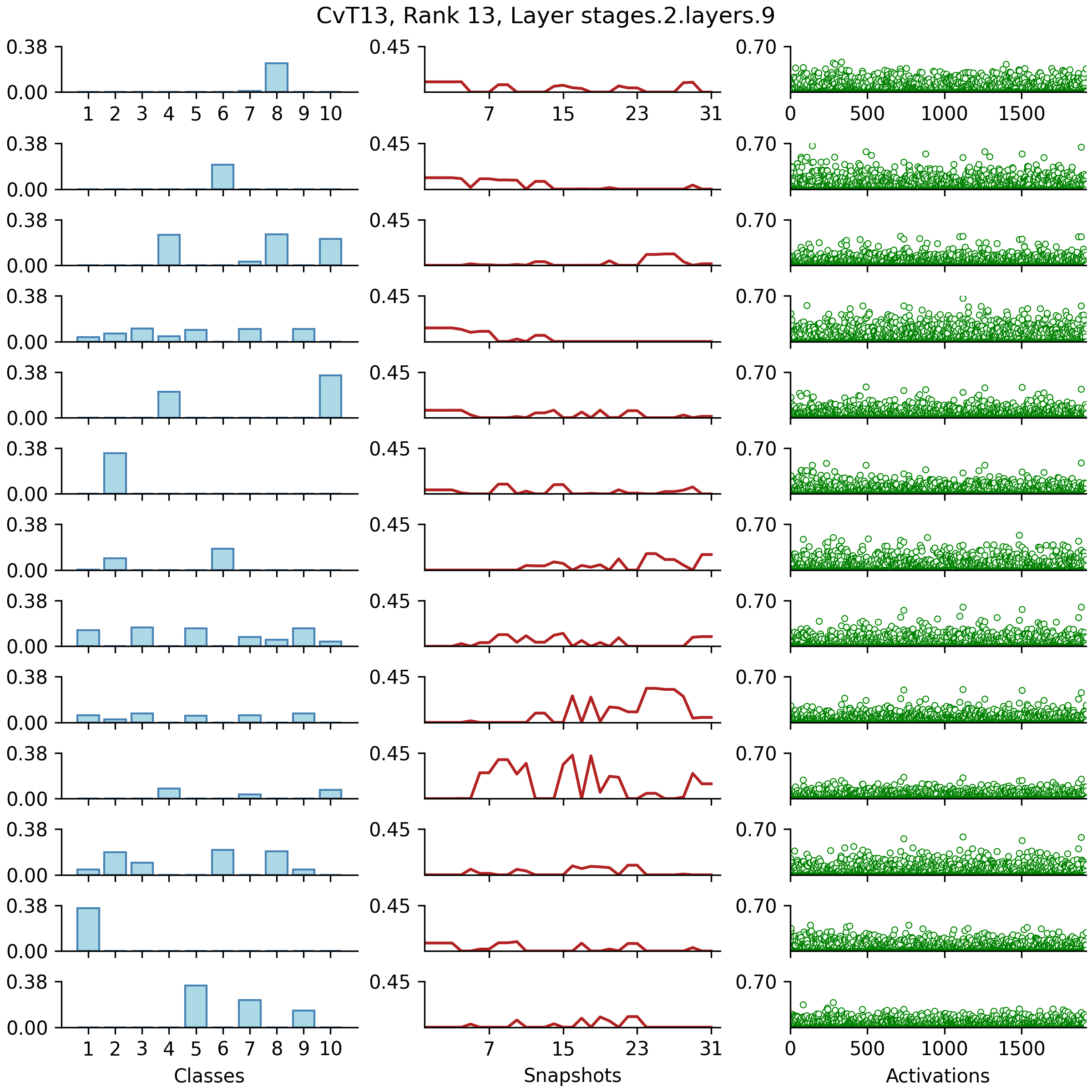}
\caption{Example of tensor component analysis on an activations tensor: two rank-13 TCA plots of activation tensors from CvT13 trained on Split-CIFAR-10, permuted to order factors that are best aligned. The similarity score between the two is 0.49. \textit{Left}-TCA plot of activations from the Naive strategy, with a reconstruction error of 0.94. \textit{Right}-TCA plot of activation from the Replay strategy, with a reconstruction error of 0.87.}
\label{fig:tca_naive_replay_cifar10}
\end{figure*}

\begin{figure*}[!htbp]
\centering
\includegraphics[width=0.48\textwidth]{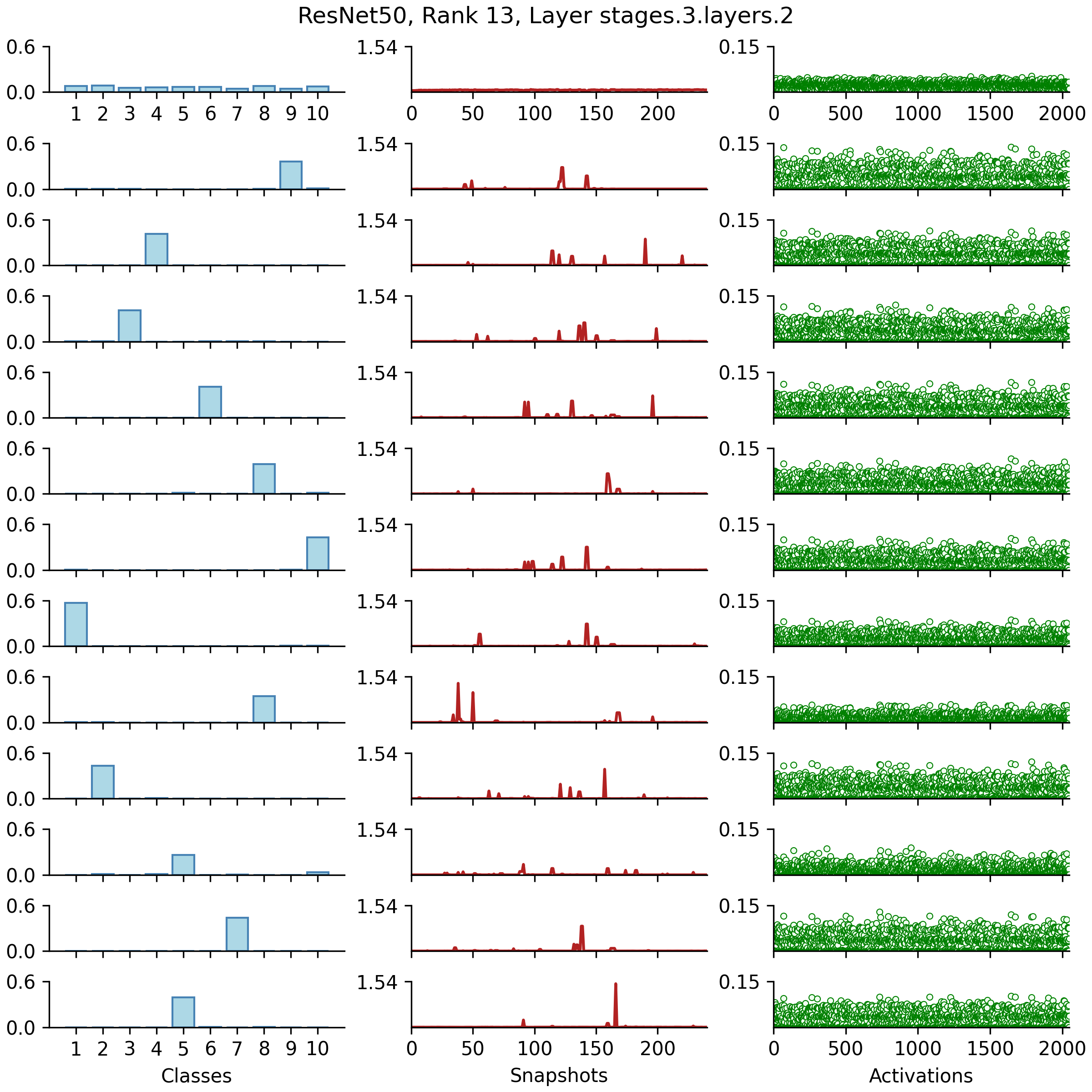}
\includegraphics[width=0.48\textwidth]{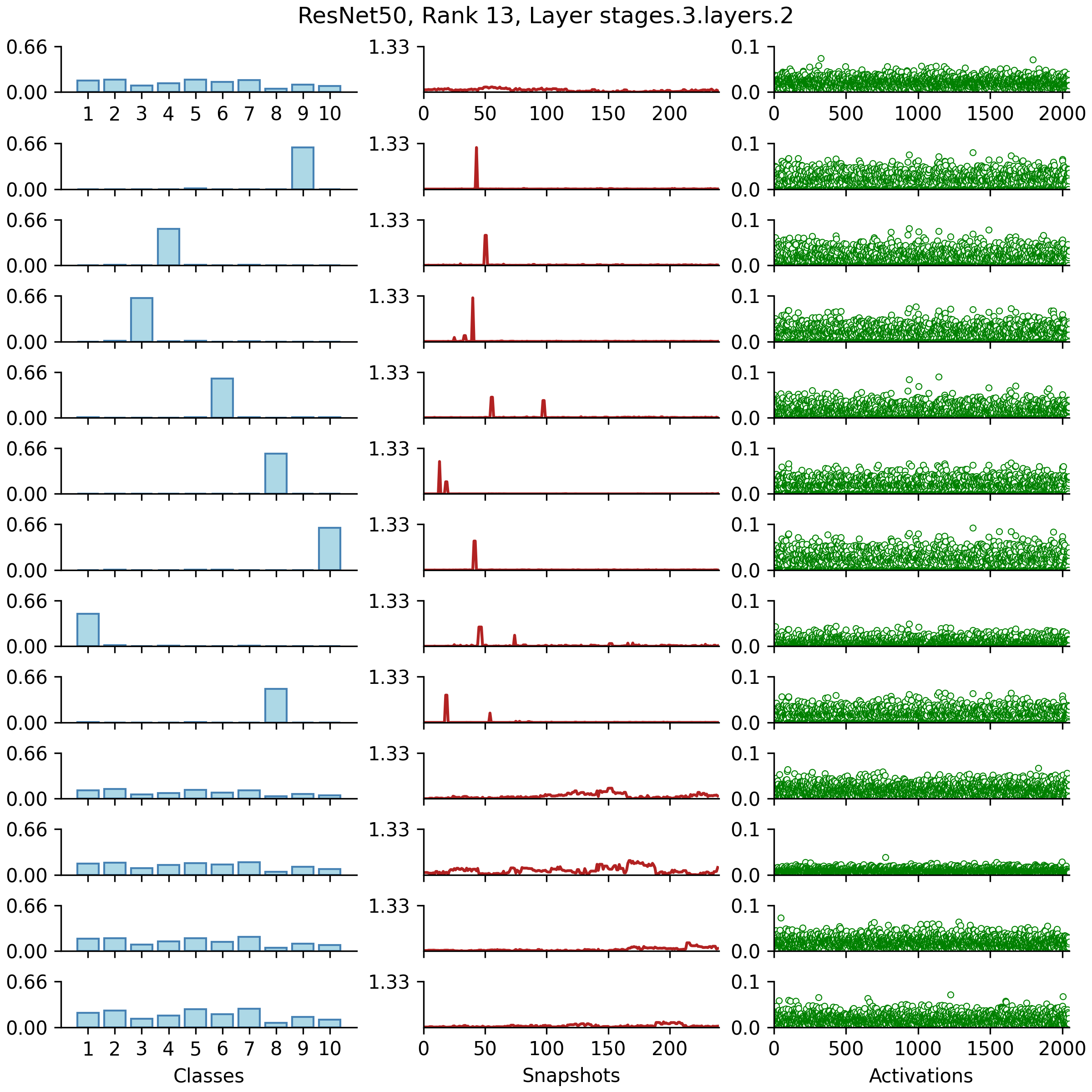}
\caption{Example of tensor component analysis on an activations tensor: two rank-13 TCA plots of activation tensors from ResNet50 trained on Split-CIFAR-100, permuted to order factors that are best aligned. The similarity score between the two is 0.53. \textit{Left}-TCA plot of activations from the MAS strategy, with a reconstruction error of 0.17. Out of the 100 classes, we selected 10, where each class belongs to a distinct task. \textit{Right}-TCA plot of activation from the WSN strategy, with a reconstruction error of 0.58.}
\label{fig:tca_mas_wsn_cifar100}
\end{figure*}

\subsection{Additional Filter Tensor Decomposition Plots}
\label{sec:more_filter_tca}
\begin{figure*}[!htbp]
\centering
\includegraphics[width=0.48\textwidth]{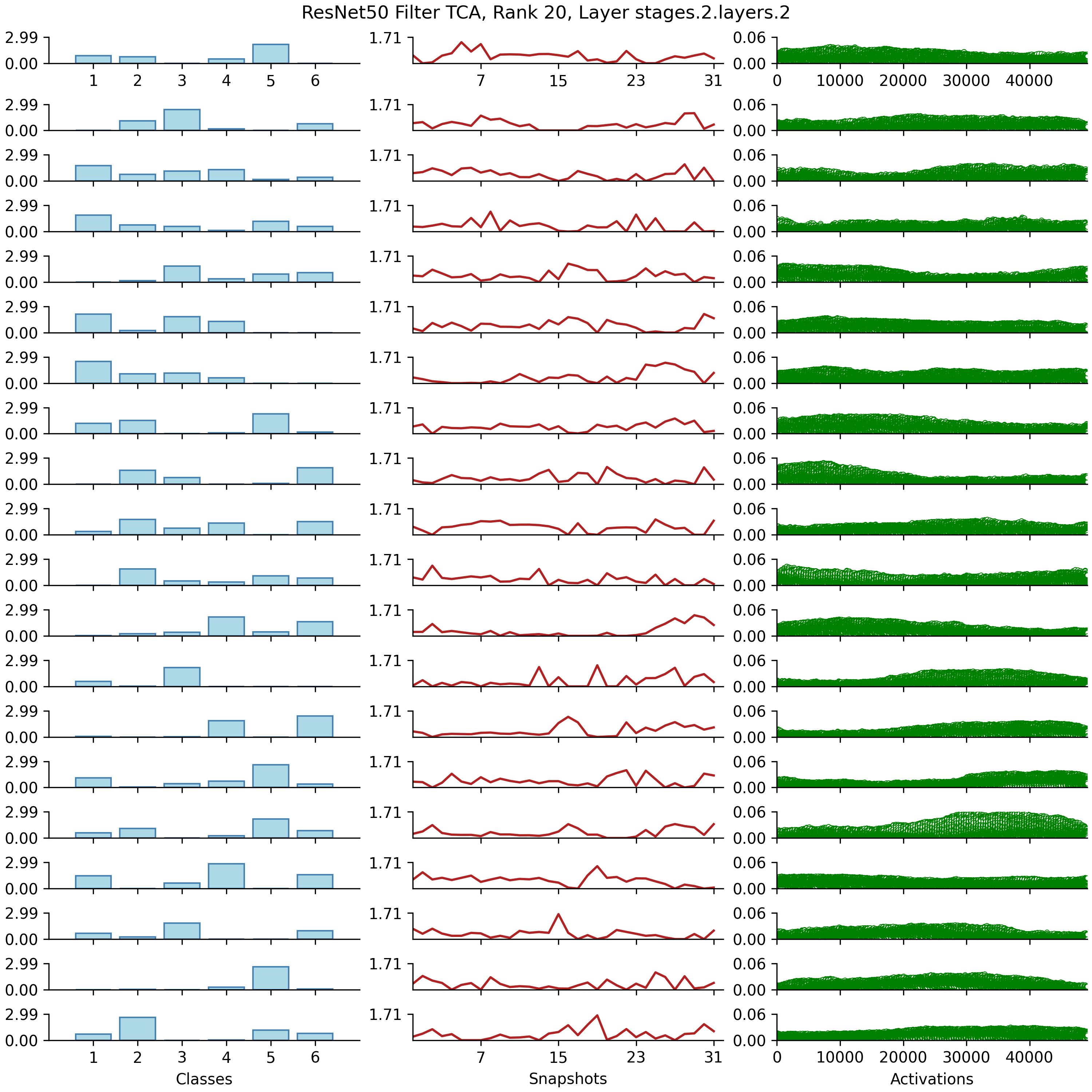}
\includegraphics[width=0.48\textwidth]{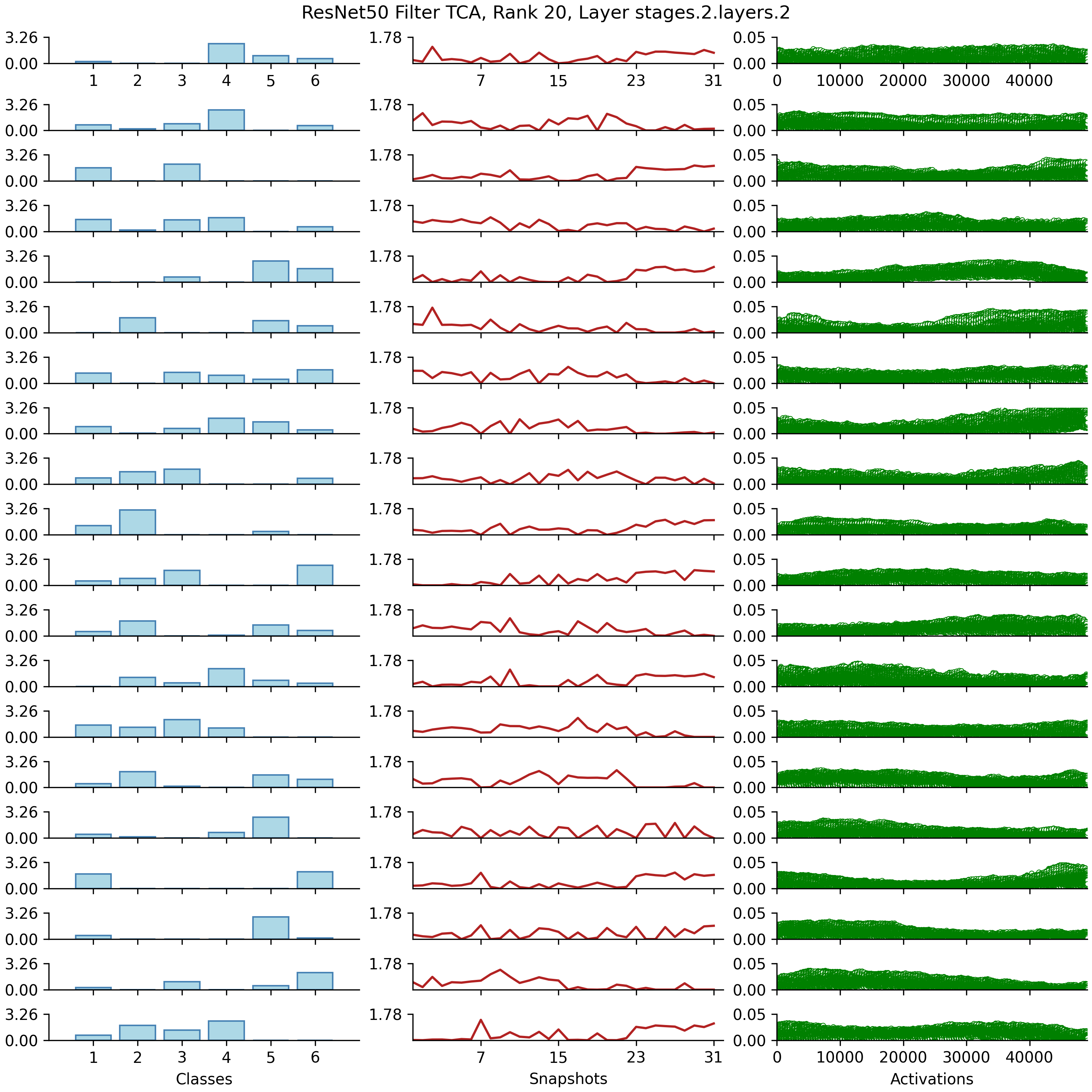}
\caption{Example of tensor component analysis on an optimized images for filters tensor: two rank-20 TCA plots of optimized images for filters tensors from ResNet50 trained on Split-CIFAR-10. \textit{Left}-TCA plot of filter images from the EWC strategy, with a reconstruction error of 0.22. \textit{Right}-TCA plot of activation from the MAS  strategy, with a reconstruction error of 0.22.}
\label{fig:tca_filter_ewc_mas}
\end{figure*}

\begin{figure*}[!htbp]
\centering
\includegraphics[width=0.48\textwidth]{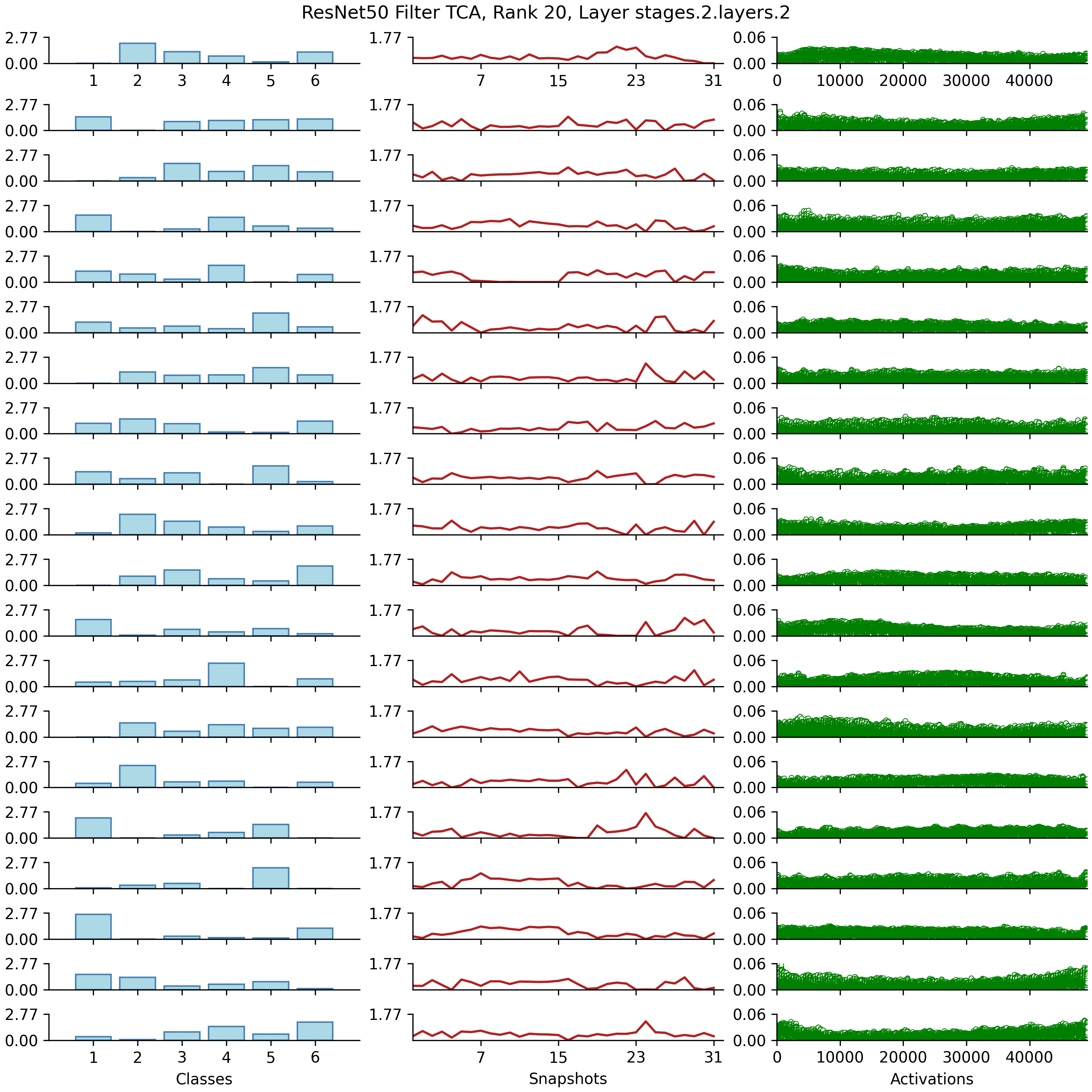}
\includegraphics[width=0.48\textwidth]{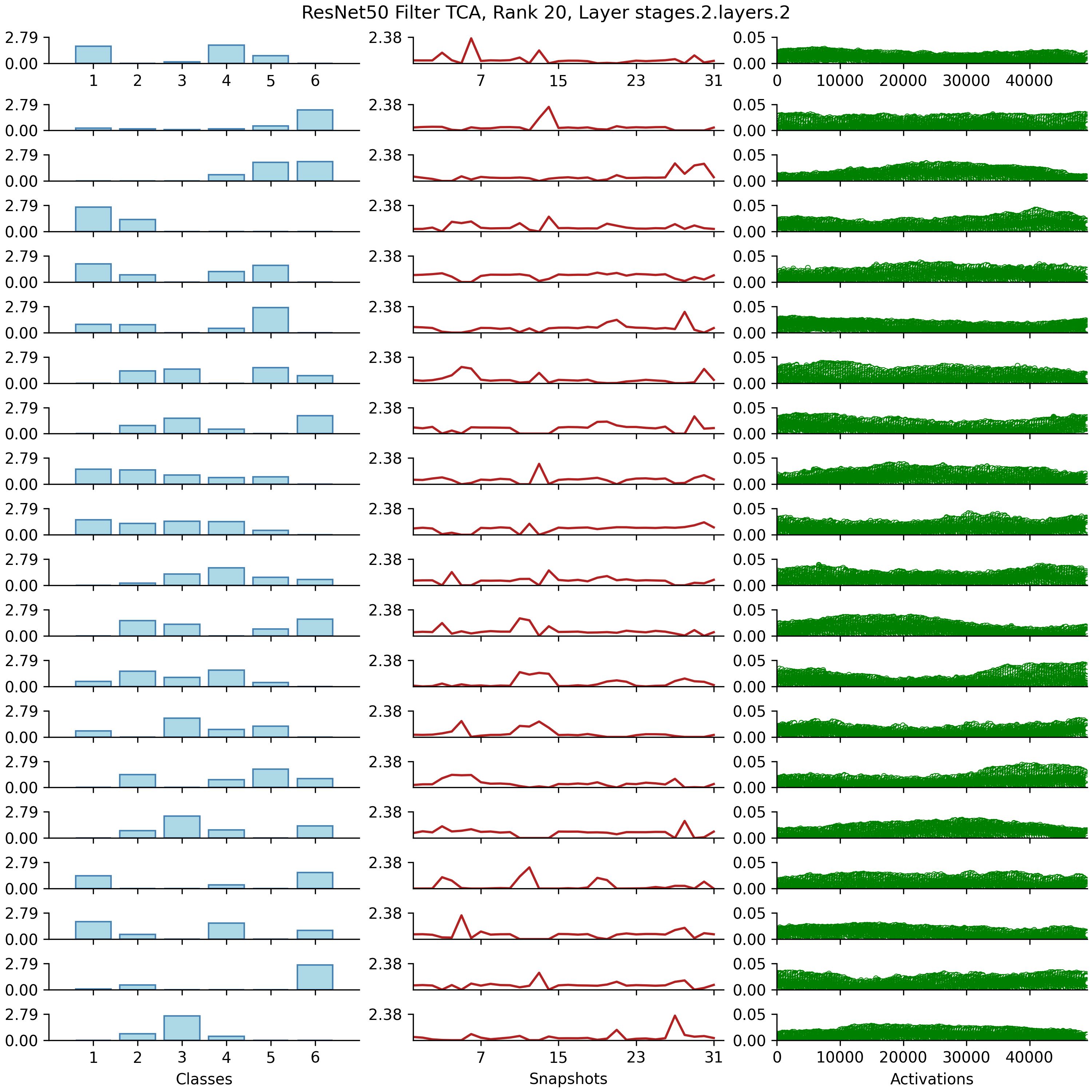}
\caption{Example of tensor component analysis on an optimized images for filters tensor: two rank-20 TCA plots of optimized images for filters tensors from ResNet50 trained on Split-CIFAR-10. \textit{Left}-TCA plot of filter images from the Replay strategy, with a reconstruction error of 0.20. \textit{Right}-TCA plot of activation from the RMN strategy, with a reconstruction error of 0.24.}
\label{fig:tca_filter_replay_rmn}
\end{figure*}

%% file: v4/supplementary_c.tex
\section{TRAINING RESULTS}
\label{sec:suppl_training_results}

\begin{figure*}[!tbp] %!htbp
% \centering
\includegraphics[width=0.945\textwidth]{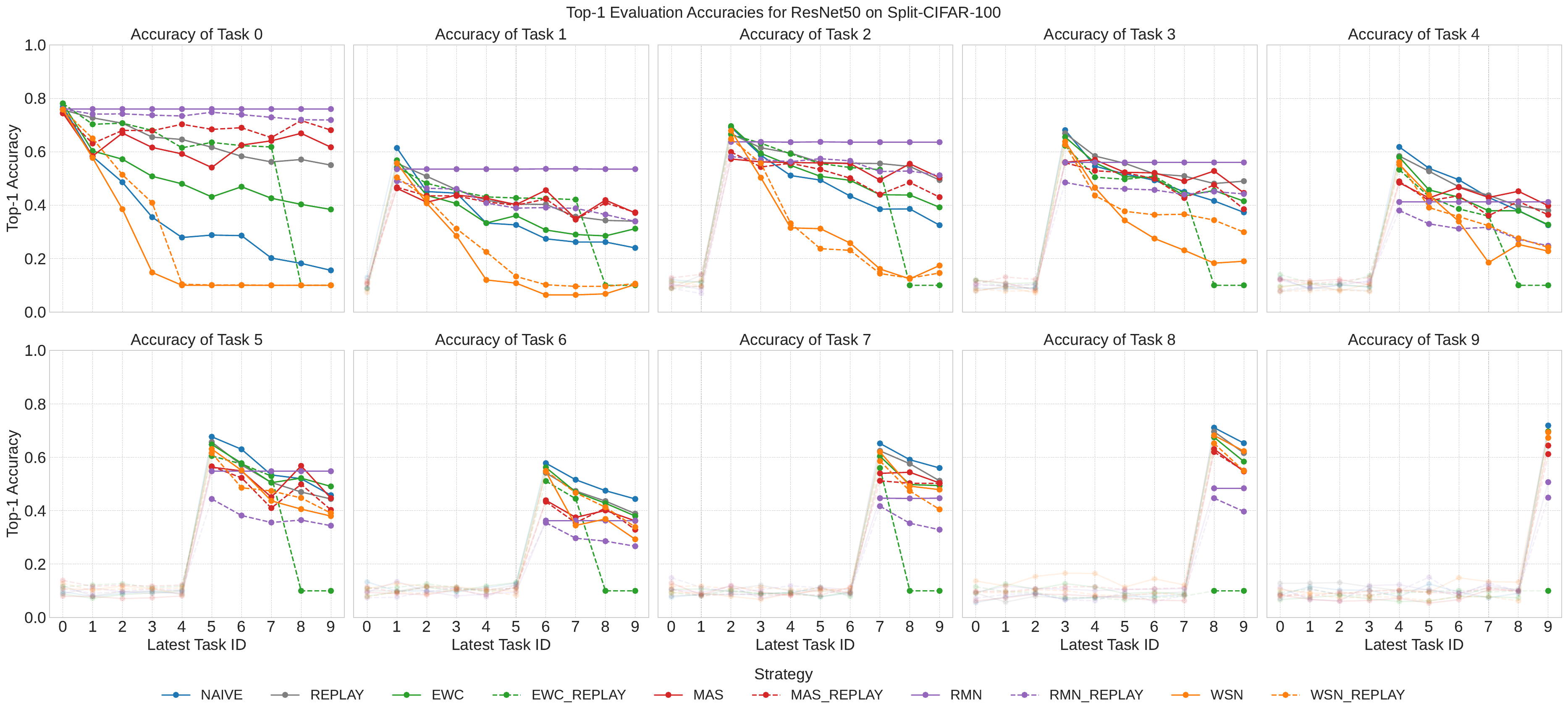}
\includegraphics[width=0.945\textwidth]{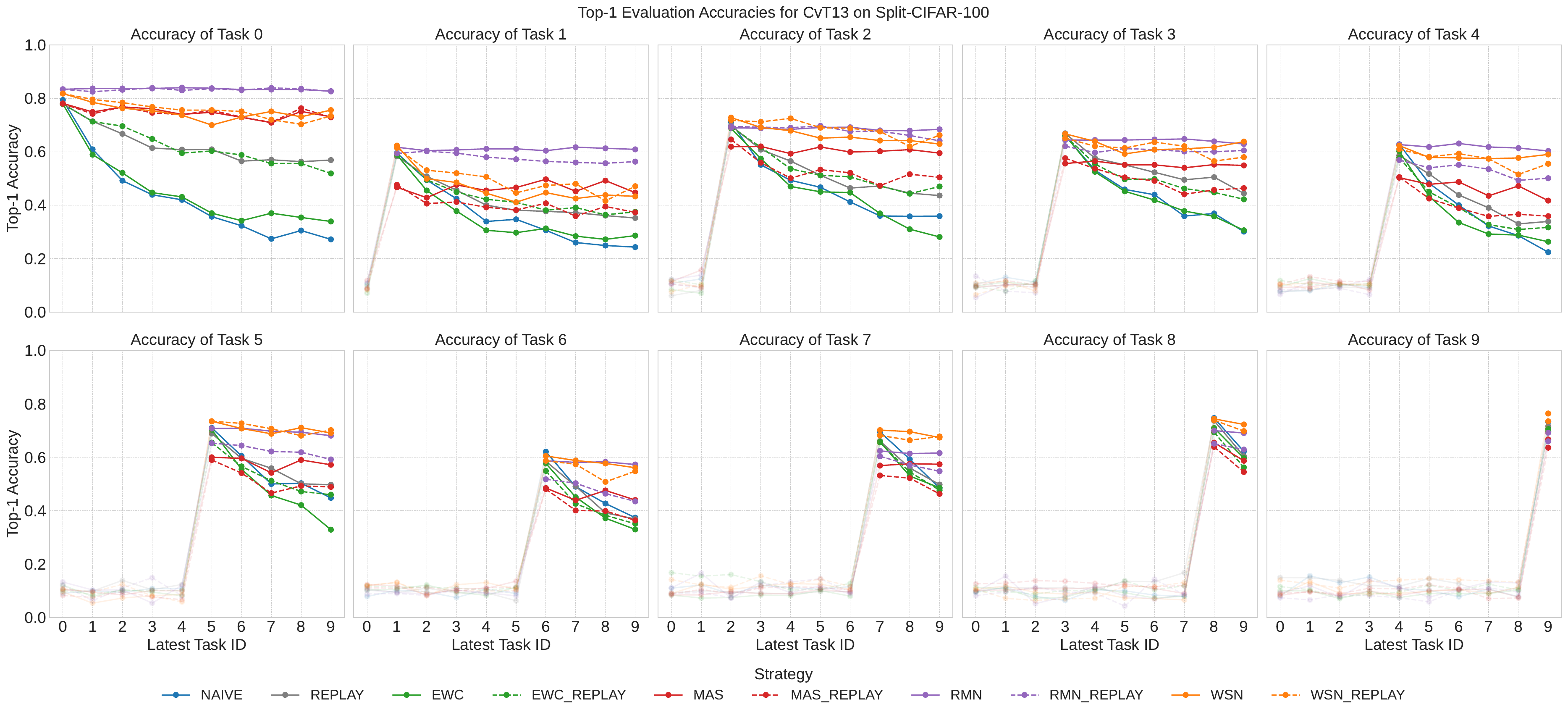}
\includegraphics[width=0.945\textwidth]{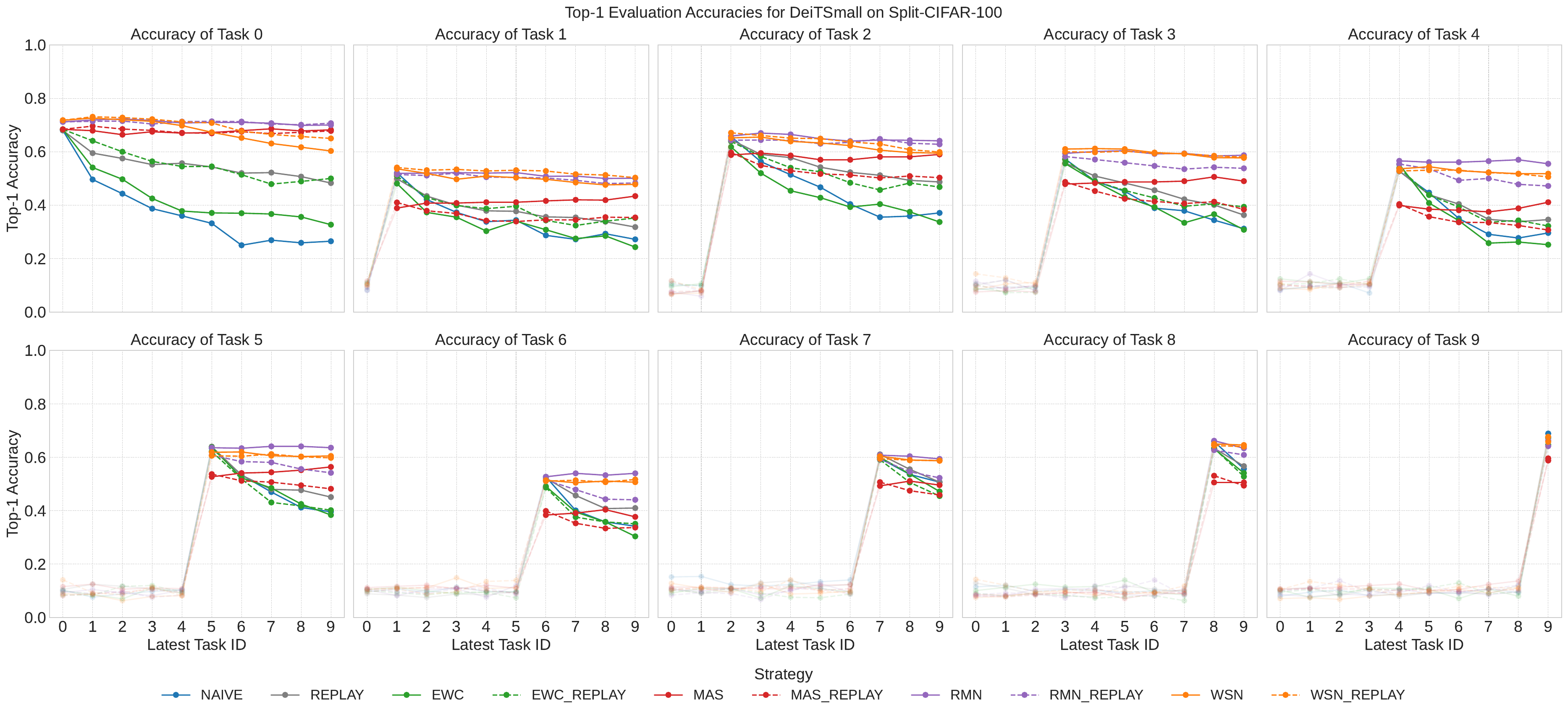}
\caption{Results on the Split-CIFAR-100 dataset. \textit{Upper Row} - ResNet50, \textit{Middle Row} - CvT13, \textit{Bottom Row} - DeiTSmall.}
\label{fig:cifar100_training}
\end{figure*}

\begin{figure*}[!t]
\centering
\includegraphics[width=0.945\textwidth]{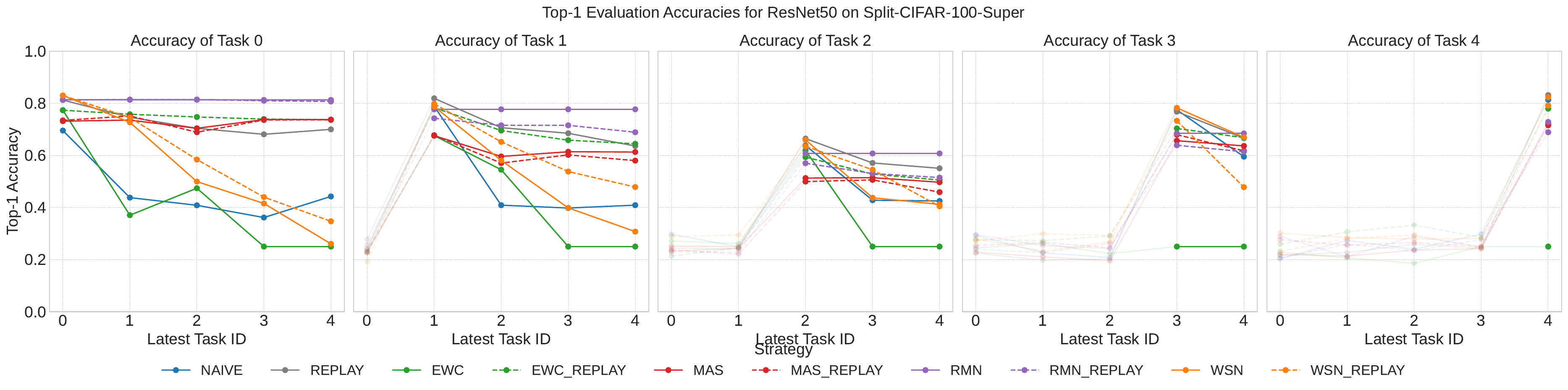}
\includegraphics[width=0.945\textwidth]{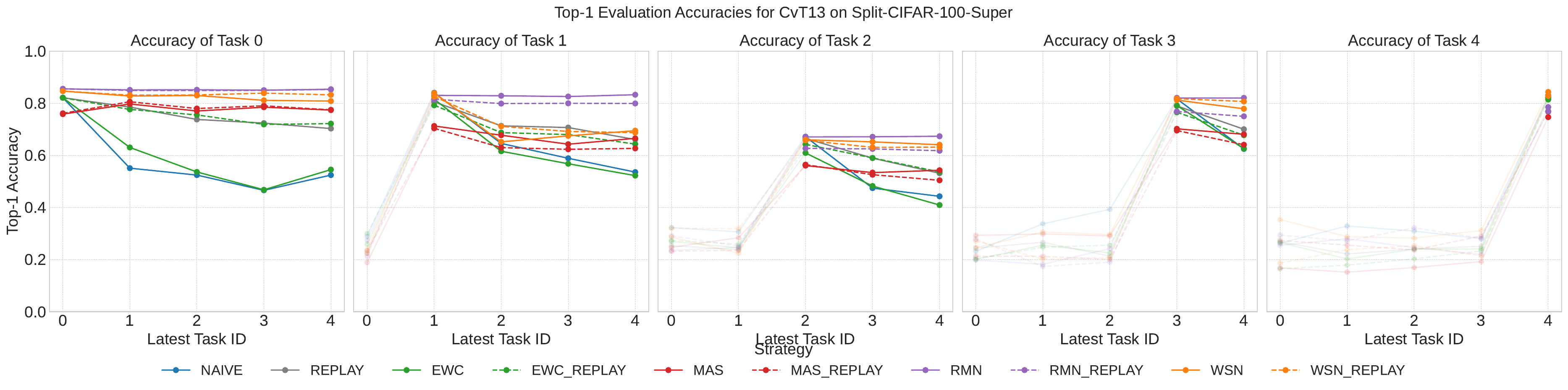}
\includegraphics[width=0.945\textwidth]{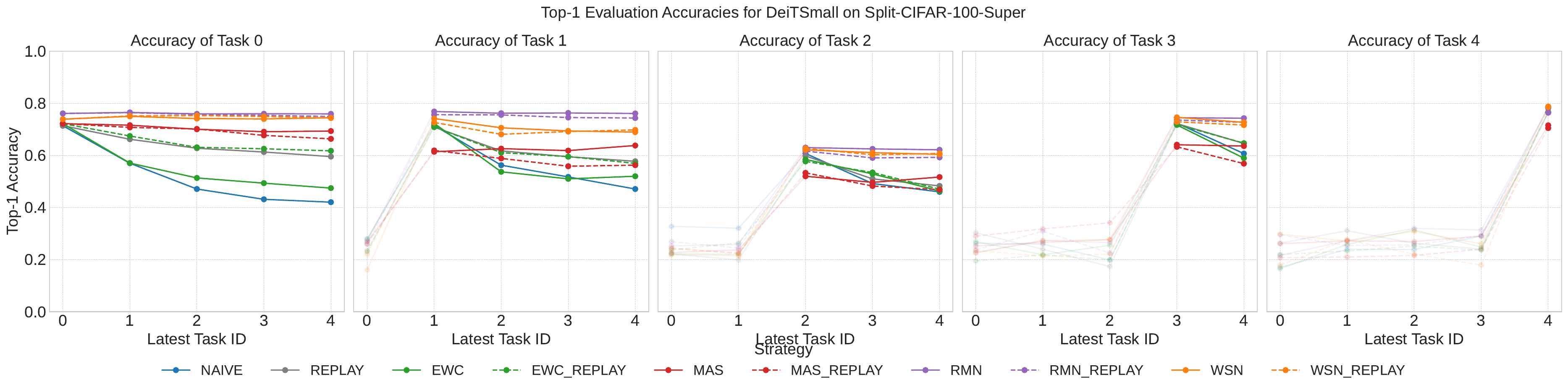}
\caption{Results on the Split-CIFAR-100-Super dataset. \textit{Upper Row} - ResNet50, \textit{Middle Row} - CvT13, \textit{Bottom Row} - DeiTSmall.}
\label{fig:cifar100super_training}
\end{figure*}

Figures \ref{fig:cifar10_training}, \ref{fig:cifar100_training}, and \ref{fig:cifar100super_training} display top-1 evaluation accuracies across different architectures and datasets using the CL strategies from Table \ref{table:strategies}. We highlight the evaluation accuracy on a task only after the model has encountered the task.

For a better visualization of our training results, we encourage readers to view the results on our dashboard at: \url{https://wandb.ai/nishantaswani/cl_decomp/reportlist}

\begin{figure*}[!tbp]
% \centering
\includegraphics[width=0.945\textwidth]{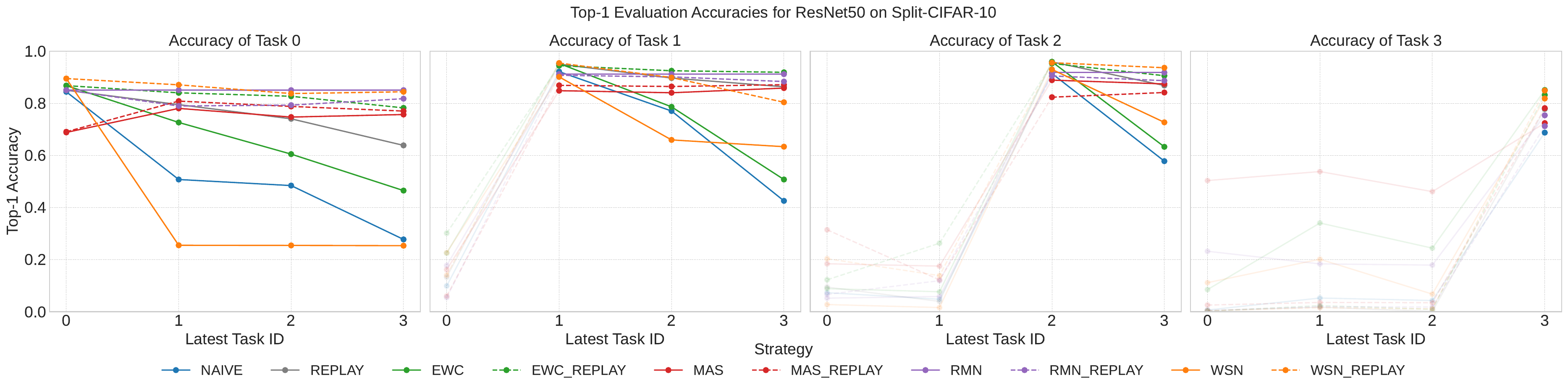}
\includegraphics[width=0.945\textwidth]{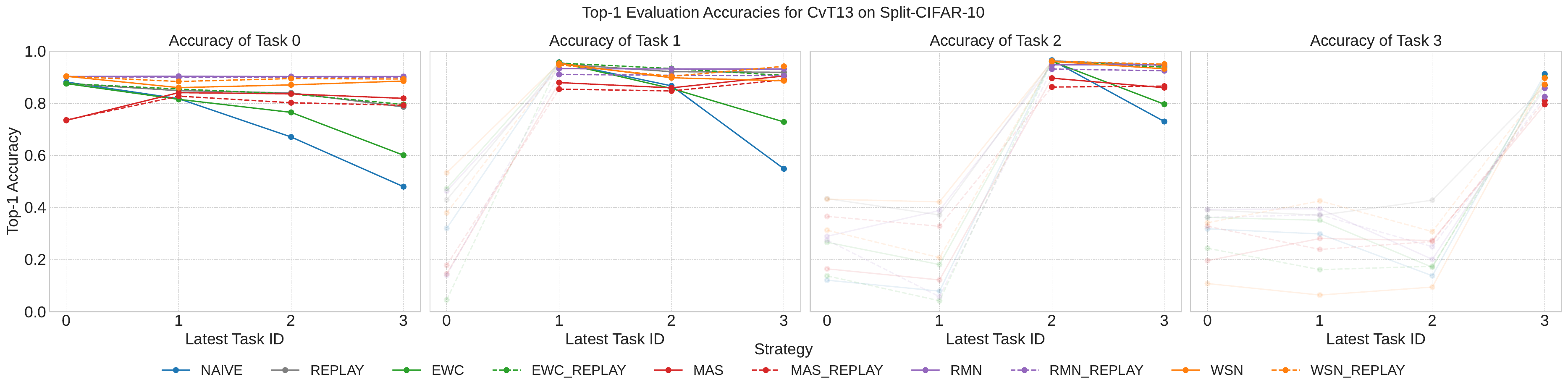}
\includegraphics[width=0.945\textwidth]{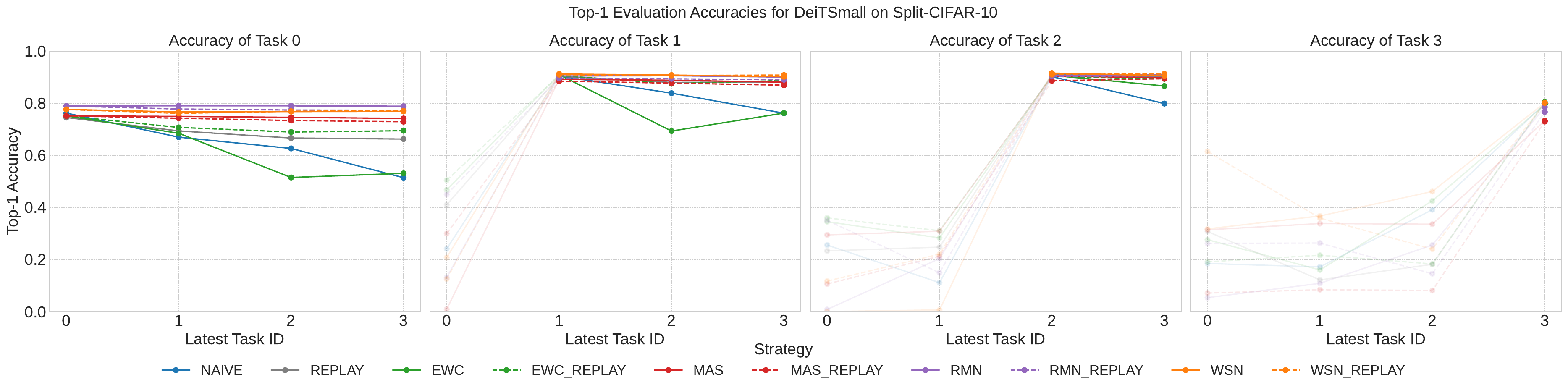}
\caption{Results on the Split-CIFAR-10 dataset. \textit{Upper Row} - ResNet50, \textit{Middle Row} - CvT13, \textit{Bottom Row} - DeiTSmall.}
\label{fig:cifar10_training}
\end{figure*}

%% file: v4/supplementary_d.tex
\section{CODE}
\label{sec:suppl_code}

All of our project code is documented and publicly available at \url{https://github.com/niniack/CLDecomp}.